\documentclass{article}

     \usepackage[nonatbib, final]{neurips_2019}

\usepackage[utf8]{inputenc} %
\usepackage[T1]{fontenc}    %
\usepackage{hyperref}       %
\usepackage{url}            %
\usepackage{booktabs}       %
\usepackage{amsfonts}       %
\usepackage{nicefrac}       %
\usepackage{microtype}      %
\usepackage{graphicx}
\usepackage{amsmath}
\usepackage{amssymb}
\usepackage{rotating}
\usepackage{enumitem}
\usepackage[dvipsnames]{xcolor}
\title{Predicting the Politics of an Image Using Webly Supervised Data}

\author{%
  Christopher Thomas \hspace{2em} Adriana Kovashka\\
  Department of Computer Science\\
  University of Pittsburgh\\
  Pittsburgh, PA 15213\\
  \texttt{\{chris,kovashka\}@cs.pitt.edu} \\
}

\begin{document}

\maketitle

\begin{abstract}
The news media shape public opinion, and often, the visual bias they contain is evident for human observers. This bias can be inferred from how different media sources portray different subjects or topics. In this paper, we model visual political bias in contemporary media sources at scale, using webly supervised data.
We collect a dataset of over one million unique images and associated news articles from left- and right-leaning news sources, and develop a method to predict the image's political leaning. 
This problem is particularly challenging because of the enormous intra-class visual and semantic diversity of our data.
We propose a two-stage method to tackle this problem. 
In the first stage, the model is forced to learn relevant visual concepts that, when joined with document embeddings computed from articles paired with the images, enable the model to predict bias. In the second stage, we remove the requirement of the text domain and train a visual classifier from the features of the former model.
We show this two-stage approach facilitates learning and
outperforms several strong baselines. We also present extensive qualitative results demonstrating the nuances of the data.

\end{abstract}
\section{Introduction}

One of the goals of the media is to inform, but in practice, the media also shapes opinions \cite{happer2013role,philo2008active,angermeyer2001reinforcing,gilens1996race,schill2012visual,munoz2017image}. The same issue can be presented from multiple perspectives, both in terms of the text written in an article, and the visual content chosen to illustrate the article. 
For example, when speaking of immigration, left-leaning sources might showcase the struggles of well-meaning immigrants, while right-leaning sources might portray the misdeeds of criminal immigrants. 
The type of topics portrayed is also strong cue for the left or right bias of the source media (e.g. tradition is primarily seen as a value on the right, while diversity is seen as a value on the left \cite{atlantic}). 

In this paper, we present 
a method for recognizing the political bias of an image, which we define as whether the image came from a left- or right-leaning media source. 
This requires understanding: 1) what visual concepts to look for in images, and 2) how these visual concepts are portrayed across the spectrum. 
Note that this is a very challenging task because many of the concepts that we aim to learn show serious visual variability within the left and right. 
For example, the concept of ``immigration'' can be illustrated with a photo of a border wall, children crying behind bars while detained, immigration agents, protests and demonstrations about the issue, politicians giving speeches, etc. 
Human viewers account for such within-class variance by generalizing what they see into broader semantic concepts or themes using prior knowledge, deduction, and reasoning.

On the other hand, modern CNN architectures learn by discovering recurring textures or edges representing objects in the images through backpropagation. 
However, the same objects might appear and be discussed \emph{across} the political spectrum, meaning that the simple presence or absence of objects is not a good indicator of the politics of an image. 
Thus, model training may fall into poor local minima due to the lack of a recurring discriminative signal. Further, it is not merely the presence or absence of objects that matters, but rather \textit{how} they are portrayed, often in subtle ways.

In order to capture the visual concepts necessary to predict the politics of an image, we propose a method which uses an auxiliary channel at training time, namely the article text that the image is paired with. 
Our method contains two stages. 
In the first one, we learn a document embedding model on the articles, then train a model to predict the bias of the image, given the image and the paired document embedding. To be successful on this task, the model learns to recognize visual cues which complement the textual embedding and suggest the politics of the image-text pair. 
At test time, we want to recognize bias from images alone, without any article text. Thus, in the second training stage of the model, we use the first stage model as a feature extractor and train a linear bias classifier on top. The article text serves as a type of privileged information to help guide learning.

Since recognizing the right semantic and visual concepts amidst intra-class variance requires large amounts of data, we train our approach on webly supervised data: the only labels are in the form of the political leaning of the source that the image came from. However, for testing purposes, we collect human annotations and test on %
images where annotators agreed on the label. 
We experimentally show that our method outperforms numerous baselines on both a large held-out webly supervised test set, and the set of crowdsourced annotations.

We believe that recognizing the political bias of a photograph is an important step towards building socially-aware computer vision systems. Such awareness is necessary if we hope to use computer vision systems to automatically tag or describe images (e.g. for the visually impaired) or to summarize  large collections of potentially biased visual content. Social media companies or search engines may deploy such techniques to automatically identify the political bent of images or even entire news sites being spread or linked to. Progress has already been made in this space in other domains. For example, Facebook automatically determines users' political leanings from site activity and pages liked \cite{merrill_2016}. Other works have studied predicting political affiliation from text \cite{conover2011predicting, wong2016quantifying, volkova2014inferring} or even MRI scans \cite{schreiber2013red}. However, \emph{visual} bias understanding has been greatly underexplored. While some work examines \textit{visual persuasion} \cite{joo2014visual, Hussain_2017_CVPR}, none analyzes political leaning as we do.

Our contributions are as follows:
\begin{itemize}[noitemsep,nolistsep]
    \item We propose and make available\footnote{Our dataset, code, and additional materials are available online for download here:  \href{http://www.cs.pitt.edu/~chris/politics}{\texttt{http://www.cs.pitt.edu/\(\sim \)chris/politics}}} a very large dataset of biased images with paired text, and a large amount of diverse crowdsourced annotations regarding political bias.
    \item We propose a weakly supervised method for predicting the political leaning of an image by using noisy auxiliary textual data at training time.
    \item We perform a detailed experimental analysis of our method on both webly supervised and human annotated data, and demonstrate %
    the factors humans use to predict bias in images.
    \item We show qualitative results that demonstrate the relationship between images and semantic concepts, and the variability in how faces of the same person appear on the left or the right.
\end{itemize}

\section{Related Work}

\paragraph{Weakly supervised learning.}

Our work is in the weakly supervised setting, in the sense that other than noisy left/right labels, our method does not receive information about what makes an image left- or right-leaning. This is challenging because there is significant variety in the type of content that can be left-leaning or right-leaning. Thus, our method needs to identify relevant visual concepts based on which to make its predictions. 
Recently, weakly supervised approaches have been proposed for classic topics such as object detection \cite{oquab2015object,cinbis2016weakly,zhou2016learning,wei2018ts2c, Ye_2019_ICCV}, action localization \cite{wang2017untrimmednets,richard2017weakly}, etc. 
Researchers have also developed techniques for learning from potentially noisy web data, e.g. \cite{chen2015webly}.
Also related is work in unsupervised discovery of patterns and topic modeling, e.g. \cite{li2018patternnet,li2017mining,sicre2017unsupervised,singh2012unsupervised,zhou2010unsupervised,jae2013style,doersch2012makes,sivic2005discovering,fei2005bayesian}.
In contrast to these works, our problem exhibits much larger within-class variance (with left and right being the classes of interest). Unlike objects and actions, the differences between left and right live in semantic space as much as they do in visual space, hence our use of auxiliary training inputs. 

\vspace{-0.3cm}
\paragraph{Curriculum learning.}
Also relevant are self-paced and curriculum learning approaches  \cite{jiang2015self,pentina2015curriculum,zamir2017feedback,zhang2017curriculum,jiang2018mentornet}. 
These attempt to simplify learning by finding ``easy'' examples to learn with first. 
We too employ a type of curriculum learning. We first train a multi-modal classifier to predict bias, using the assumption that the relation between text and bias is more direct. We then leverage this model as a feature extractor by adding an image-only politics classifier on top of it. Thus, our method focuses the model on relevant visual concepts using text.

\vspace{-0.3cm}
\paragraph{Privileged information.}
Our method also exploits a similar intuition as privileged information methods \cite{vapnik2015learning,sharmanska2013learning,hoffman2016learning,motiian2016information,elliott2017imagination,Gomez_2017_CVPR,borghi2018face,Lambert_2018_CVPR} that use an extra feature input at training time. These approaches use tied weights \cite{borghi2018face}, computing summary statistics \cite{sharmanska2013learning, Lambert_2018_CVPR}, or multitask training \cite{elliott2017imagination} to guide learning. The closest such method to ours is \cite{Gomez_2017_CVPR} which uses an approach trained to predict text embeddings from images. The features are then applied on visual-only data. However, in early experiments we showed directly predicting text embeddings from images is much more challenging on our data because of the many-to-many relationship of images with topics (e.g.~image of the White House can be paired with text about Trump's children, border control, LGBT rights, etc.).

\vspace{-0.3cm}
\paragraph{Connecting images and text.}
To learn the meaning of the images, we elevate the image representation to a semantic one, by connecting images and text. However, because our texts contain a lot of information not relevant to the image, our main method does not predict text from the image. 
The latter task has received sustained interest 
\cite{vinyals2015show,Donahue_2015_CVPR,Johnson_2016_CVPR,Venugopalan_2017_CVPR,Pedersoli_2017_ICCV,show_adapt_tell,DaiICCV17,Anderson_2018_CVPR,Eisenschtat_2017_CVPR,yepami2019} 
but our domain is unique in that articles that are paired with our images are orders of magnitude longer.

\vspace{-0.3cm}
\paragraph{Visual rhetoric.}

Our work also belongs to a recent trend of developing algorithms to analyze visual media and the strategies that a media creator uses to convey a message. \cite{joo2014visual,joo2015automated} analyze the skills and characteristics that a politician is implied to have through a photo, e.g. ``competent''; we adapt their method as a baseline in our setting. \cite{peng2018same} study differences in facial portrayals between presidential candidates, and \cite{wang2017polarized,wang2016deciphering} examine visual differences between supporters of the left or right.
We learn to \textit{generate} faces from the left and right. 
Further, we examine differences in general images rather than just faces. 
\cite{Hussain_2017_CVPR,yepami2019} predict the persuasive messages of advertisements, but persuasion in political images is more subtle.
These works are based on careful and expensive human annotations, while we aim to discover facets of bias in a weakly supervised way.

\vspace{-0.3cm}
\paragraph{Bias prediction in language.}

Prior work in NLP has discovered indicators of biased language and political framing (i.e. presenting an event or person in a positive or negative light). For example, \cite{recasens2013linguistic,baumer2015testing} use carefully designed dictionary, lexical, grammatical and content features to detect biased language, using supervision over short phrases. Others \cite{pennacchiotti2011machine,cohen2013classifying,colleoni2014echo,conover2011predicting, wong2016quantifying, volkova2014inferring} have studied predicting politics from text. In contrast, 
it is not clear what ``lexicon'' of biased content to use for images. %

\section{Dataset}
\label{sec:data}

Because no dataset exists for this problem, we assembled a large dataset of images and text about contemporary politically charged topics.
We got a list of ``biased'' sources from \verb|mediabiasfactcheck.com| which places news media on a spectrum from extreme left to extreme right.
We used \cite{isidewith} to get a list of current ``hot topics'' e.g. immigration, LGBT rights, welfare, terrorism, the environment, etc.  We crawled the media sources that were labeled left/right or extreme left/right for images using each of these topics as queries. After identifying images associated with each keyword and the pages they were on, we used \cite{peters2013content} to extract articles. We obtained 1,861,336 images total and 1,559,004 articles total. We manually removed boilerplate text (headers, copyrights, etc.) which leaked into some articles. 

\subsection{Data deduplication}
\label{sec:data:dedup}

Because sources cover the same events, some images are published multiple times. To prevent models from ``cheating'' by memorization, all experiments are performed on a ``deduplicated'' subset of our data. We extract features from a Resnet \cite{he2016deep} model for all images. Because computing distances between all pairs is intractable, we use \cite{malkov2016efficient} for approximate \textit{k}NN search ($k=200$). We set a threshold on neighbors' distances to find duplicates and near-duplicates. We determine the threshold empirically by examining hundreds of \textit{k}NN matches to ensure all near-duplicates are detected. From each set of duplicates, we select one image (and its associated article) to remain in our ``deduplicated'' dataset while excluding all others. If the same image appeared in both left and right media sources, we keep it on the side where it was more common, e.g.~one left source and three right sources would result in preserving one of the image-text pairs from the right sources. 
After removing duplicates, we are left with 1,079,588 unique images and paired text on which the remainder of this paper is based.

\begin{figure}[t]
    \centering
    \includegraphics[width=1\linewidth]{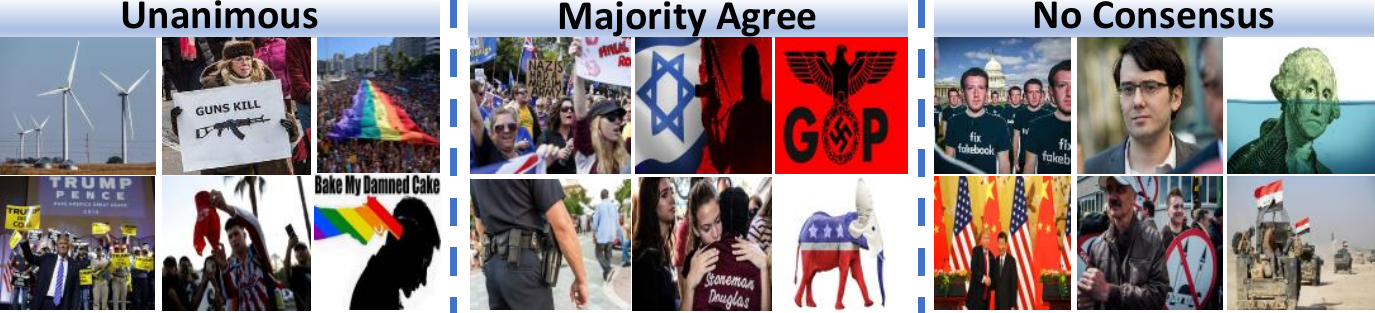}
    \vspace{-2em}
    \caption{We asked workers to predict the political leaning of images. We show examples here where  all annotators agree, the majority agree, and where there was no consensus. 
    }
    \label{fig:dataset}
    \vspace{-1.5em}
\end{figure}

\subsection{Crowdsourcing annotations}
\label{sec:annot}
\label{sec:human_concepts}

We treat the problem of predicting bias as a weakly supervised task. For training, we assume all image-text pairs have the political leaning of the source they come from. In Sec.~\ref{sec:mturk_res} we show that this assumption is reasonable by leveraging human labels, though it is certainly not correct for all images / text, e.g.~a left-leaning source may publish a right-leaning image to critique it. In order to better explore this assumption and understand human conceptions of bias, we ran a large-scale crowdsourcing study on Amazon Mechanical Turk (MTurk). We asked workers to guess the political leaning of images by indicating whether the image favored the left, right, or was unclear. In total, we showed 3,237 images to at least three workers each. We show examples of different levels of agreement in Fig.~\ref{fig:dataset}. In total, 993 were labeled with a clear L/R label by at least a majority. We also asked what image features
were used to make their guess. The features workers could choose (and the count of each agreed upon) was: closeup-90 (closeup of specific person's face), known person-409 (portrays public figure in political way), multiple people-237 (group or class of people portrayed in political way), no people-81 (scenes or objects associated with parties, e.g.~windmill/left, gun/right), symbols-104 (e.g.~swastika, pride flag), non-photographic-130 (cartoons, charts, etc.), logos-77 (logo of e.g.~CNN, FOX, etc.), and text in image-267 (e.g.~text on protest signs, captions, etc.).

We also showed workers the image's article and asked a series of questions about the image-text pair, such as the political leaning of the \textit{pair} (as opposed to image only), the topic (e.g.~terrorism, LGBT) the pair is related to, and which article text best aligned with the image. We computed agreement scores and found that 2.45 out 3 annotators agreed on bias label on average, while 1.71 out of 3 agreed on topic, on average. Finally, we asked workers to provide a free-form text explanation of their politics prediction for a small number of images.
We extracted semantic concepts from these explanations and later use them to train one of our baseline methods (Sec.~\ref{sec:baselines}). Humans often mentioned using the positive/negative portrayal of public figures and the gender, race and ethnicity of photo subjects. We provide a demonstration of differences in portrayal across L/R in Sec.~\ref{sec:qualitative}. Absent these cues, workers used stereotypical notions of what issues the left/right discuss or their values. For example, for images of protests or college women, annotators might guess ``left''.

To ensure quality, we used validation images with obvious bias to disqualify careless workers. We restricted our task to US workers who passed a qualification test, had $\geq98$\% approval rate, and who had completed $\geq$1,000 HITs. 
In total, we collected 14,327 sets of annotations (each containing image bias label, image-text pair bias label, topic, etc.) at a cost of \$4,771.
We include a number of experimental results on this human annotated set of images in Sec.~\ref{sec:mturk_res}.

\section{Approach}
\label{sec:approach}

We hypothesize that the complementary textual domain provides a useful cue to guide the training of our visual bias classifier. 
The text of the articles includes words that clearly correlate with political bias, e.g. ``unite'', ``medicaid'', ``donations'', ``homosexuality'', ``Putin'', ``Antifa'' and ``brutality'' strongly correlate with left bias according to our model, while ``defend'', ``retired'', ``NRA'', ``minister'' and ``cooperation'' strongly correlate with right bias.
By factoring out these semantic concepts into the auxiliary text domain, we enable our model to learn complementary visual cues. 
We use information flowing from the visual pipeline, and fuse it with the document embedding as an auxiliary source of information. Because we are primarily interested in \textit{visual} political bias, we next remove our model's reliance on textual features, but keep all convolutional layers fixed. We train a linear bias classifier on top of the first model, using it as a feature extractor. Thus, at \emph{test time,} our model predicts the bias of an image \textit{without using any text}.
We illustrate our method in Fig.~\ref{fig:method}.

\subsection{Method details}
\label{sec:method_details}

\begin{figure}[t]
    \vspace{-0.5em}
    \centering
    \includegraphics[width=1\linewidth]{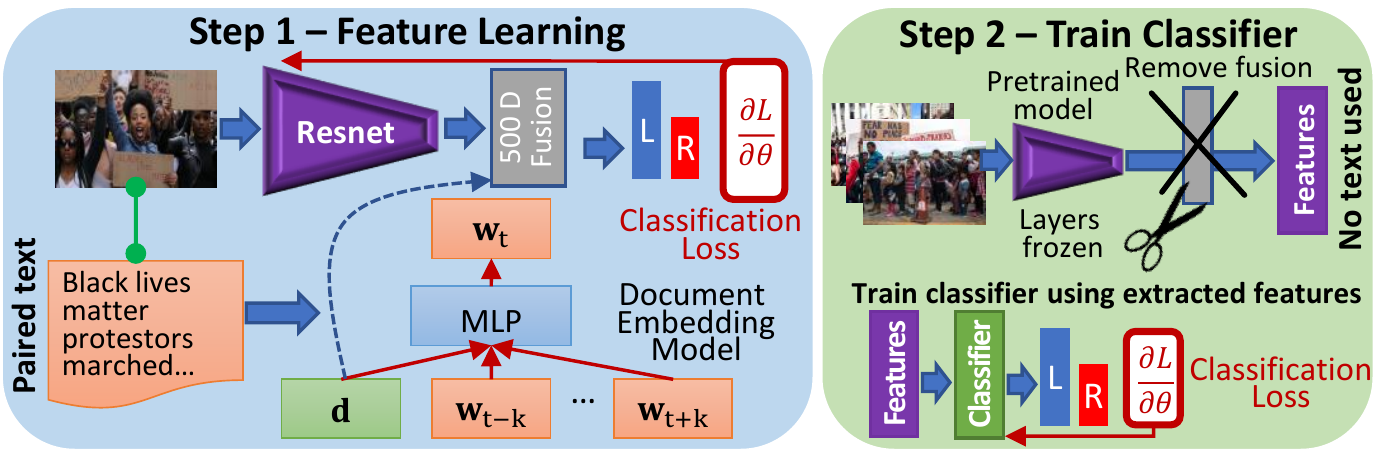}
    \vspace{-2em}
    \caption{We propose a two-stage approach. In stage 1, we learn visual features jointly with paired text for bias classification. In stage 2, we remove the text dependency by training a classifier on top of our prior model using purely visual features. 
    We show that this approach significantly outperforms directly training a model to predict bias. See Sec.~\ref{sec:method_details} for details. %
    }
    \label{fig:method}
    \vspace{-1em}
\end{figure}

We wish to capture the implicit semantics of an image by leveraging the association between images and text. More specifically, let
\vspace{-0.3cm}
\begin{equation}
    \mathcal{D} = \left\{ \mathbf{x}_i, \mathbf{a}_i, \mathbf{y}_i \right\}_{i=1}^{N}
\end{equation}
denote our dataset $\mathcal{D}$, where $\mathbf{x}_i$ represents image $i$, $\mathbf{a}_i$, represents the textual article associated with the $i^{th}$ image, and $\mathbf{y}_i$ represents the political leaning of the image. In the first stage of our method, we seek the following function: 
\begin{equation}
    f_{\mathbf{\theta}}\left(\mathbf{x}_i, \Omega\left(\mathbf{a}_i\right)\right) = \mathbf{y}_i
\end{equation}
where $\Omega\left(.\right)$ represents transforming the article text into a latent feature space. 
We train Doc2Vec \cite{le2014distributed} offline on our train set of articles to parameterize $\Omega$.
Specifically, $\Omega$ is trained to maximize the average log probability
\vspace{-0.2cm}
\begin{equation}
\label{eq:doc2vec_train}
    \frac{1}{T} \sum_{t=1}^{T} %
    \log p\left(\mathbf{w_{t}} | \mathbf{d}, \mathbf{w_{t-k}}, \ldots, \mathbf{w_{t+k}} \right) 
\end{equation}
where $T$ is the number of words in article $\mathbf{a}$ (we omit the index $i$ to simplify notation), $p$ represents the probability of the indicated word, $\mathbf{w_{t}}$ is the learned embedding for word $t$ of article $\mathbf{a}$, $\mathbf{d}$ is the learned document embedding of $\mathbf{a}$ (200D), and $k$ is the window around the word to look when training the model. We use hierarchical softmax \cite{morin2005hierarchical} to compute $p$. 
We train Doc2Vec on our corpus of news articles, and observe more intuitive embeddings than from a pretrained model.

After training, we compute $\Omega$ for a given article $\mathbf{a}$ by finding the embedding $\mathbf{d}$ that maximizes Eq.~\ref{eq:doc2vec_train}.
$\Omega$ thus projects each article into a space where the resulting vector captures the overall latent context and topic of the article.
We provide $\Omega\left(\mathbf{a}\right)$ to our model's fusion layer for each train image. The fusion layer is a linear layer which receives concatenated image and text features and learns to project them into a multimodal image-text embedding space which is finally used by the classifier.

The formulation of $f_\theta(.)$ described above requires that the \textit{ground-truth} text be available at test time and also does not ensure that our model is learning \textit{visual} bias (i.e.~the classifier may be relying primarily on text features and ignoring the visual channel completely).
To address this problem, in the second stage of our method, we finetune $f_\theta$ to directly predict the politics of an \emph{image only}, without the text, as follows: $f^\prime_{\mathbf{\theta}, \theta^{\prime}}\left( \mathbf{x}_i \right) = \mathbf{y}_i$. Specifically, we freeze the trained convolutional parameters of $f_\theta$ and add a final linear classifier layer to the network, whose parameters are denoted $\theta^\prime$. Because $f_\theta$'s convolutional layers have already been trained jointly with text features, they have 
already learned to extract visual features which complemented the text domain; we now learn to use those features \textit{alone} for bias prediction, as shown in Fig.~\ref{fig:method}.

\subsection{Implementation details}
\label{sec:impl}
All methods use the Resnet-50 \cite{he2016deep} architecture and are initialized with a pretrained Imagenet model. We train all models using Adam \cite{kingma2014adam}, with learning rate of 1.0e-4 and minibatch size of 64 images. We use cross-entropy loss and apply class-weight balancing to correct for slight data imbalance between L/R. We use an image size of 224x224 and random horizontal flipping as data augmentation. We use Xavier initialization ~\cite{glorot2010understanding} for non-pretrained layers. We use PyTorch \cite{paszke2017automatic} to train all image models. For our text embedding, we use \cite{rehurek_lrec}, with $\mathbf{d} \in \mathcal{R}^{200\times1}$ and train using distributed memory \cite{le2014distributed} for 20 epochs with window size $k=20$, ignoring words which appear less than 20 times.

\section{Experiments}
\label{sec:experiments}

In this section, we demonstrate our method's performance at predicting left/right bias.
We show results on a large held-out test set from our dataset, whose left/right labels come from the leaning of the news source containing the image.
We also show results on test images for which a majority of human annotators agreed on the bias and show how humans reason about visual bias. %
We show that seeing the complementary text information helped \textit{humans} become more accurate at this task, much like seeing the text at training time helps our algorithm. 
We also show the challenge of our task through across-class nearest-neighbors, how the portrayal of politicians differs from the left to the right, images that best match various words from articles, and visualize how our model makes decisions about visual bias. Our supp.~contains additional results such as results per-media source / per-political issue, an exploration of the learned text embedding space, failure cases for machines/humans, humans' reasoning behind their bias decisions, and examples from our dataset.

\subsection{Methods compared}
\label{sec:baselines}

For quantitative results, we show the accuracy of each method on predicting left/right bias.
We compare against the following baselines:
\begin{itemize}[noitemsep,nolistsep,leftmargin=*]
    \item \textsc{Resnet \cite{he2016deep}} - A standard 50-layer classification Resnet. %
    \item \textsc{Joo \cite{joo2014visual}} - Adaptation of Joo et al.'s method for our task. 
    We use \cite{joo2014visual}'s dataset to train predictors for 15 attributes and nine ``intents'' (qualities the photo subject is estimated to have, e.g. trustworthiness, competence). We then use the predictions for these attributes and intents on images from our dataset as additional features to a Resnet to predict a left/right leaning.
    \item \textsc{Human Concepts} - We use the manually extracted vocabulary of bias-related concepts (e.g. ``confederate'', ``African-American'') from the human-provided explanations (Sec.~\ref{sec:annot}) and download data for each from Google Image Search. We train a separate Resnet to predict concepts, and use it on each image in our dataset: $p(c_j|\mathbf{x}_i)$ denotes the probability that image $\mathbf{x}_i$ exhibits concept $c_j$. We then use the confidence of each detected concept, as a feature vector to predict bias. %
    \item \textsc{OCR} - We use \cite{Liao2018Text} to recognize free-form scene text in images. Because images contain words not found in the default lexicon (e.g.~Manafort), we create our own lexicon from the 100k most common words in our articles. We use \cite{symspell} for spelling correction. %
    We represent each recognized word as its learned word embedding,
    denoted $\mathbf{w}^\prime_i$, weighed by the confidence of the recognition $p\left(\mathbf{w}^\prime_i\right)$ as provided by the recognition model. The feature is thus given by $\frac{1}{n} \sum_{i=1}^{n} p\left(\mathbf{w}^\prime_i\right) \mathbf{w}^\prime_i$.
\end{itemize}

All methods use the same residual network architecture. For methods relying on additional features, we use the fusion architecture in Fig.~\ref{fig:method}. 
For reference, we also show an upper-bound method  \textsc{Ours (GT)} which uses the \textbf{G}round \textbf{T}ruth text paired with the images \textit{at test time} (to compute a document embedding), in addition to the image. We thus consider it an upper-bound to the task of visual only prediction. \textsc{Ours (GT)} is the same as the first stage of our approach (see Fig.~\ref{fig:method}, left), without the addition of the image classifier layer in step 2.

\subsection{Evaluating on weakly supervised labels}
\label{sec:main_res}

In Table \ref{tab:results}, we show the results of evaluating our methods on 75,148 held-out images with weakly supervised labels.
Our method performs best overall. The top two performing methods rely on semantics discovered in the text domain (\textsc{Ours} and \textsc{OCR}). \textsc{OCR} is unique in that it is able to explicitly use text information at test time, by discovering text within the image and then using word embeddings. 
\textsc{Ours} improves over \textsc{OCR} by 2.6\% (relative 3.8\%, reduction in error of 8\%).
The improvement of \textsc{Ours} over \textsc{Resnet} is 3.4\% (relative 5\%, error reduction of 11\%). 
This amounts to classifying an additional {\raise.17ex\hbox{$\scriptstyle\mathtt{\sim}$}}2,555 images correctly.
Relying on the concepts humans identified actually slightly \textit{hurt} performance compared to  \textsc{Resnet}. This may be because of a disconnect between humans' preconceived notions about L/R %
and those required by the dataset.
We finally observe \textsc{Joo} performs the weakest, likely because \cite{joo2014visual}'s data mainly features closeups of politicians, while ours contains a much broader image range.

\begin{table}[t]
    \vspace{-0.5em}
    \centering
    \begin{tabular}{|c||c|c|c|c||c||c|c|}
    \hline
    \textbf{Method} & \textbf{\textsc{Resnet}} & \textbf{\textsc{Joo}} & \textbf{\textsc{Human Concepts}} & \textbf{\textsc{OCR}} &  \textbf{\textsc{Ours}} &  \textbf{\textsc{Ours (GT)}} \\ \hline \hline
    \textbf{Accuracy} & 0.678 & 0.670 & 0.675 & 0.686 & \textbf{0.712} & 0.803 \\ \hline
    \end{tabular}
    \caption{Accuracy on weakly supervised labels with the best visual-only prediction method in bold.}
    \label{tab:results}
    \vspace{-.75em}
\end{table}

\subsection{Evaluating on human labels}
\label{sec:mturk_res}

We next tested our methods on test images which at least a majority of MTurkers labeled as having the same bias, i.e.~those that humans agreed had a particular label. We describe this dataset in Sec.~\ref{sec:annot}. Because workers also labeled images with what features of the image they used to make their prediction, we also break down each method's performance by feature. We show this result in Table \ref{tab:mturk_result}. \textsc{Ours} performs best on average across all categories and performs best on four out of eight categories. Categories where \textsc{Ours} is outperformed on are reasonable: \textsc{OCR} performs best when text can be relied on in the image, i.e.~``logos'' and ``text in image''. We note that while the overall result for \textsc{OCR} approaches \textsc{Ours}, \textsc{Ours} works better on a broader set of images than \textsc{OCR} and is thus a more general method for predicting \textit{visual} bias. \textsc{Ours} is also outperformed by \textsc{Human Concepts} when humans relied on a known face (politician, celebrity, etc.). This may be because \textsc{Human Concepts} relies on external training data (Sec.~\ref{sec:baselines}) which feature many known individuals, e.g.~``rappers'' and ``founding fathers''. \textsc{Joo} outperforms our method when the prediction depends on scene context (``no people''), again likely because \textsc{Joo} uses an external human-labeled dataset to learn features, including scene attributes (e.g.~indoor, background, national flag, etc.). We note \textsc{Ours (GT)} performs sig.~worse on human labels vs.~weakly-supervised labels. This is likely because \textsc{Ours (GT)} has learned to exploit dataset-specific features (e.g.~author names, header text, etc.) for prediction, which does not actually translate into humans' commonsense understanding of political bias.

\begin{table}[t]
    \vspace{-0.36em}
    \centering
    \resizebox{1\columnwidth}{!}{
    \begin{tabular}{|c||c|c|c|c||c||c|}
    \hline
    \textbf{Feature/Method} & \textbf{\textsc{Resnet}} & \textbf{\textsc{Joo}} & \textbf{\textsc{Human Concepts}} &  \textbf{\textsc{OCR}} &  \textbf{\textsc{Ours}} & \textbf{\textsc{Ours (GT)}}  \\ \hline \hline
    \textbf{Closeup} & 0.567 & 0.544 & 0.622 & 0.578 & \textbf{0.656} & 0.578 \\ \hline
    \textbf{Known Person} & 0.567 &  0.550 & \textbf{0.570} & 0.560 & 0.521 & 0.575 \\ \hline
    \textbf{Multiple People} & 0.722 & 0.671 & 0.688 & 0.730 &  \textbf{0.768} & 0.705 \\ \hline
    \textbf{No People} & 0.556 & \textbf{0.605} & 0.494 & 0.580 & 0.593 & 0.667 \\ \hline
    \textbf{Symbols} & 0.558 & 0.596 & 0.548 & 0.577 & \textbf{0.606} & 0.587 \\ \hline
    \textbf{Non-Photographic} & 0.577 & 0.569 & 0.584 & 0.577 &  \textbf{0.585} & 0.654 \\ \hline
    \textbf{Logos} & 0.545 & 0.584 & 0.597 & \textbf{0.662} & 0.623 & 0.584 \\ \hline
    \textbf{Text in Image} & 0.629 & 0.625 & 0.596 & \textbf{0.637} &  0.607 & 0.659 \\ \hline \hline
    \textbf{Average} & 0.590 & 0.593 & 0.587 & 0.613 &  \textbf{0.620} & 0.626 \\ \hline 
    \end{tabular}
    }
    \caption{Accuracy on human consensus labels with the best visual-only prediction method in bold. }
    \label{tab:mturk_result}
    \vspace{-2em}
\end{table}

We next test whether our assumption that all images harvested from a right- or left-leaning source exhibit that type of bias is reasonable. Several results computed from our ground-truth human study suggest that our web labels are a reasonable approximation of bias. 
First, we observe that the relative performance of the methods across Table \ref{tab:results} and \ref{tab:mturk_result} is roughly maintained; \textsc{Ours} is best, followed by \textsc{OCR}, and the other methods essentially tied. 
The results are also sound, e.g.~when humans used text, \textsc{OCR} tends to do better, which indicates the model's concept of bias correlates with humans'.

We also performed two other experiments to verify our conclusions. First, we explored the difference between the performance of our method on images on which the \textit{majority} of humans agreed vs.~those on which humans \textit{unanimously} agreed. We found that our method worked better when humans unanimously labeled the images vs.~simple majority (gain of 4.4\%). This suggests that as humans become more certain of bias, our model (trained on noisy data) also performs better. Next, we evaluated the impact of text on humans' bias predictions. We compared how humans \textit{changed} their predictions (made originally using the image only) after they saw the text paired with the image. 
We found that when workers picked a L/R label, the label was strongly correlated with the weakly supervised label. Moreover, after seeing the text, humans became even more correct with respect to the noisy labels, switching many ``unclear'' predictions to the ``correct'' label (i.e.~the noisy label). This indicates that: 1) our noisy labels are a good approximation of the true bias of the images; and 2) the paired text is useful for predicting bias (a result also borne out by our experiments).

\subsection{Quantitative ablations}
\label{sec:ablation}
In order to test the soundness of our method and our experimental design, we performed several ablations. We first tested the importance of the second stage of our method (right side of Fig.~\ref{fig:method}). To do so, we used \textsc{Ours (GT)}, the result of the first stage of our method and
instead of performing stage 2, we removed the dependency on text by zeroing out all text embedding weights in the fusion layer.
We evaluated on our weakly supervised test set and obtained 0.677, a result sig.~worse than our full method, underscoring the importance of stage 2. We next tested how the performance of our method varied given the length of the article text. 
We thus trained our method with the first \textit{k} sentences of the article and obtained these results: $k=1\rightarrow0.672$, $k=2\rightarrow0.669$, $k=5\rightarrow0.668$, $k=10\rightarrow0.669$. All choices of \textit{k} tested performed sig.~worse than using the full article (0.712).

We finally examined how reliant our method was on images from a particular media source being in our train set (i.e.~to test if the model was learning non-generalizable, source-specific features). 
We experimented with leaving out all training data harvested from a few popular sources. The result was (before excluding~$\rightarrow$ after excluding): Breitbart (0.607$\rightarrow$0.566),  CNN (0.873$\rightarrow$0.866),  CommonDreams (0.647$\rightarrow$0.636),  DailyCaller (0.703$\rightarrow$0.667),   DemocraticUnderground (0.713$\rightarrow$0.700), NewsMax (0.685$\rightarrow$0.628), and TheBlaze (0.746$\rightarrow$0.742). We observed only a slight decrease for all sources we tested, suggesting our method is not dependent on seeing the source at train time.

\vspace{-1em}
\subsection{Qualitative results} 
\label{sec:qualitative}

\begin{figure*}[t]
    \vspace{-0.5em}
    \centering
    \includegraphics[width=1\textwidth]{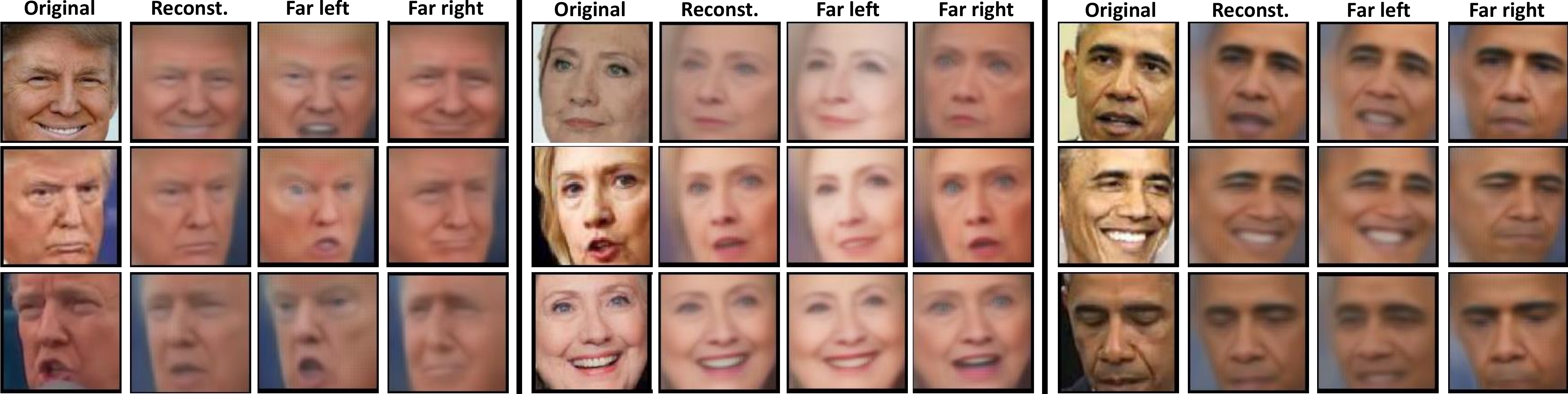}
    \vspace{-0.5cm}
    \caption{We modified photos to be more left/right.
    We show the model's ``reconstruction'' of each face next to the original sample, followed by the sample transformed to the far left and right.}
    \label{fig:trump}  
    \vspace{-0.3em}
\end{figure*}

\vspace{-0.1cm}
\paragraph{Modeling facial differences across politics: } Many workers noted how politicians were portrayed in making their decision (Sec.~\ref{sec:human_concepts}). 
To visualize the differences in how well-known individuals are portrayed within our dataset, we trained a generative model to modify a given Trump/Clinton/Obama face, and make it appear as if it came from a left/right leaning source. We use a variation of the autoencoder-based model from \cite{chris_bmvc2018}, which learns a distribution of facial attributes and latent features on ads, not political images.
We train the model using the features from the original method on faces of Trump/Clinton/Obama detected in our dataset using \cite{dlib09}. We use \cite{schroff2015facenet} for face recognition. To modify an image, we condition the generator on the image's embedding and modify the distribution of attributes/expressions for the image to match that person's average portrayal on the left/right, following \cite{chris_bmvc2018}'s technique. We show the results in Fig.~\ref{fig:trump}. Observe that Trump and Clinton appear angry on the far-left/right (respectively) end of the spectrum. In contrast, all three appear happy/benevolent in sources supporting their own party. We also observe Clinton appears younger in far-left sources. In far-right sources, Obama appears confused or embarrassed.
These results further underscore that our weakly supervised labels are accurate enough to extract a meaningful signal.

\begin{figure}[t]
    \centering
    \vspace{-1em}
    \includegraphics[width=1\linewidth]{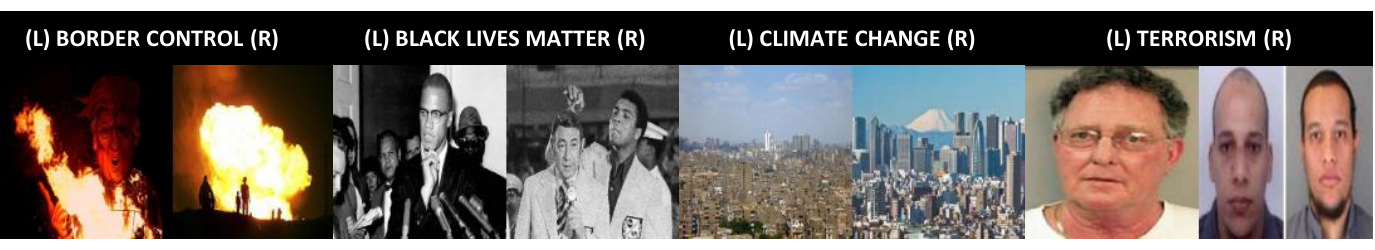}
    \vspace{-0.7cm}
    \caption{For a set of topics (e.g. LGBT, climate change), we show the closest pair of images across the left/right divide. In each pair, the image on the left is from a left-leaning source, and the one on the right is from a right-leaning source. Note how similar the images in each pair are on the surface.}
    \label{fig:closest}
    \vspace{-1em}
\end{figure}

\vspace{-0.3cm}
\paragraph{Nearest neighbors across issues and politics:}
In Fig.~\ref{fig:closest}, we show the challenge of classifying in visual space only. We compute the distance between images from the left and right, and show L/R pairs that have a small distance in feature space within topics.
For BLM, the left image is serious,
while the right image is whimsical. For climate change, one presents a more negative vision, while the other is picturesque. Both border control images show fire, but the left one is of a Trump effigy. For terrorism, the left image shows a white domestic terrorist while 
the right shows Middle-Eastern men. These pairs highlight how subtle the distinctions between L/R are for some images.

\begin{figure}[t]
    \centering
    \vspace{-0.5em}
    \includegraphics[width=1\linewidth]{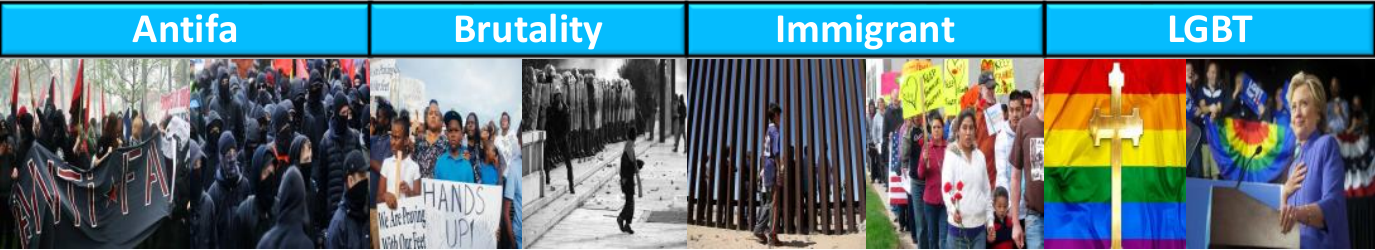}
    \vspace{-1.5em}
    \caption{We train a model to predict words from images. The model learns relevant visual cues for each word, demonstrating the utility of exploiting text, even for purely visual classification. }
    \label{fig:word_prediction}
    \vspace{-1.25em}
\end{figure}

\vspace{-0.3cm}
\paragraph{Visualizing image-text alignment:}  
We wanted to see how well our model could align images and concepts from text. We formulated a variation of our method which, instead of predicting bias, predicted relevant words. We chose a set of 1k words that had the lowest average distance between their images' features (i.e.~were visually consistent on avg.) from the 10k most frequent words.
The model is trained to predict whether each word is/is not present in the image's article given the image and text embedding. In Fig.~\ref{fig:word_prediction}, we show examples of images that were among the top-100 strongest predictions for that word. We see that the model strongly predicts ``antifa'' for black-clad protestors, ``brutality'' for police scenes and protests, ``immigrant'' for the border wall and Hispanics, and ``LGBT'' for pride flags. 
Though the image may only relate to a small portion of the lengthy text, there is enough visual signal present for the model to learn, demonstrating the benefit of leveraging text to complement the model's training.

\begin{figure}[t]
    \centering
    \vspace{.4em}
    \includegraphics[width=1\linewidth]{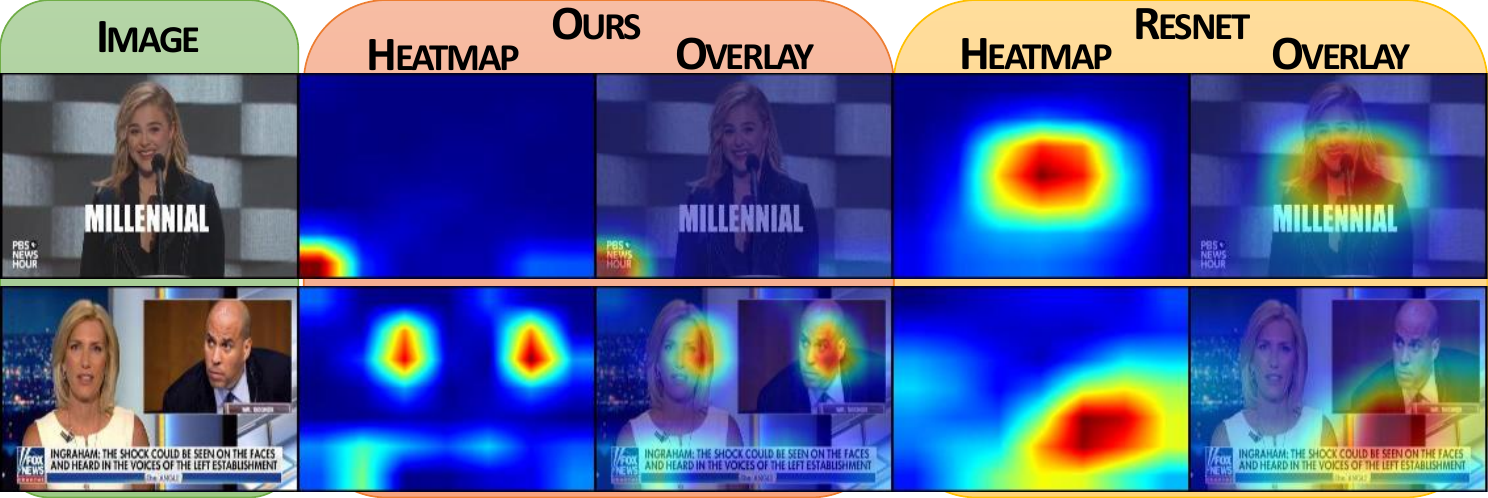}
    \vspace{-1.5em}
    \caption{We show visual explanations using \cite{chattopadhay2018grad}. We note that our model looks to logos and faces of public figures, while the baseline uses objects (e.g.~mic.) and scene type (e.g.~city in background).}
    \label{fig:gradcam}
    \vspace{-1.25em}
\end{figure}

\vspace{-0.3cm}
\paragraph{Visual explanations:} We wanted to see whether we could interpret how our model learned to perform bias classification. We used Grad-CAM++ \cite{chattopadhay2018grad} to compute attention maps on images that humans annotated. 
We show the result in Fig.~\ref{fig:gradcam}. We observe that our model pays the most attention to logos and faces of public figures. We see the model only focuses on the ``PBS'' logo in the first row (and ignores the face of the lesser known person), but pays attention to both the ``Fox News'' logo and the face of the well-known commentator in the second row. 
We believe that because our model was trained with the topic information provided via the text embedding during stage one, the visual component of the model learned to focus on learning visual features that complemented the text (such as logos and faces). Ultimately these features work better even without the text. 

\vspace{-0.7em}
\section{Conclusion}
\vspace{-0.35em}

We assembled a large dataset of biased images and paired articles and presented a weakly supervised approach for inferring the political bias of images.
Our method leverages the image's paired text to guide the model's training process towards relevant semantics in a way which ultimately improves bias classification. 
We demonstrate the contribution of our method and dataset both quantitatively and qualitatively, including on a large crowdsourced dataset.
Use cases of our work include: inferring the bias of new media sources, constructing balanced ``news feeds,'' or detecting political ads. 
Broadly speaking, our method demonstrates the potential of using an auxiliary semantic space, e.g. for abstract tasks such as video summarization and visual commonsense reasoning.

{\footnotesize \textbf{Acknowledgement:} This material is based upon work supported by the National Science Foundation under Grant Number 1566270. It was also supported by an NVIDIA hardware grant. We thank the reviewers for their constructive feedback.}

{\small
\bibliographystyle{ieee}
\bibliography{neurips_2019}

\begin{thebibliography}{10}\itemsep=-1pt

\bibitem{Anderson_2018_CVPR}
P.~Anderson, X.~He, C.~Buehler, D.~Teney, M.~Johnson, S.~Gould, and L.~Zhang.
\newblock Bottom-up and top-down attention for image captioning and visual
  question answering.
\newblock In {\em Proceedings of the IEEE Conference on Computer Vision and
  Pattern Recognition (CVPR)}, June 2018.

\bibitem{angermeyer2001reinforcing}
M.~C. Angermeyer and B.~Schulze.
\newblock Reinforcing stereotypes: how the focus on forensic cases in news
  reporting may influence public attitudes towards the mentally ill.
\newblock {\em International Journal of Law and Psychiatry}, 2001.

\bibitem{baumer2015testing}
E.~Baumer, E.~Elovic, Y.~Qin, F.~Polletta, and G.~Gay.
\newblock Testing and comparing computational approaches for identifying the
  language of framing in political news.
\newblock In {\em Proceedings of the 2015 Conference of the North American
  Chapter of the Association for Computational Linguistics: Human Language
  Technologies}, pages 1472--1482, 2015.

\bibitem{borghi2018face}
G.~Borghi, S.~Pini, F.~Grazioli, R.~Vezzani, and R.~Cucchiara.
\newblock Face verification from depth using privileged information.
\newblock In {\em British Machine Vision Conference (BMVC)}. Springer, 2018.

\bibitem{chattopadhay2018grad}
A.~Chattopadhay, A.~Sarkar, P.~Howlader, and V.~N. Balasubramanian.
\newblock Grad-cam++: Generalized gradient-based visual explanations for deep
  convolutional networks.
\newblock In {\em 2018 IEEE Winter Conference on Applications of Computer
  Vision (WACV)}, pages 839--847. IEEE, 2018.

\bibitem{show_adapt_tell}
T.-H. Chen, Y.-H. Liao, C.-Y. Chuang, W.-T. Hsu, J.~Fu, and M.~Sun.
\newblock Show, adapt and tell: Adversarial training of cross-domain image
  captioner.
\newblock In {\em Proceedings of the IEEE International Conference on Computer
  Vision (ICCV)}, Oct 2017.

\bibitem{chen2015webly}
X.~Chen and A.~Gupta.
\newblock Webly supervised learning of convolutional networks.
\newblock In {\em Proceedings of the IEEE International Conference on Computer
  Vision (ICCV)}, pages 1431--1439, 2015.

\bibitem{cinbis2016weakly}
R.~G. Cinbis, J.~Verbeek, and C.~Schmid.
\newblock Weakly supervised object localization with multi-fold multiple
  instance learning.
\newblock {\em IEEE Transactions on Pattern Analysis and Machine Intelligence
  (PAMI)}, 39(1):189--203, 2016.

\bibitem{cohen2013classifying}
R.~Cohen and D.~Ruths.
\newblock Classifying political orientation on twitter: It’s not easy!
\newblock In {\em Seventh International Association for the Advancement of
  Artificial Intelligence (AAAI) Conference on Weblogs and Social Media}, 2013.

\bibitem{colleoni2014echo}
E.~Colleoni, A.~Rozza, and A.~Arvidsson.
\newblock Echo chamber or public sphere? predicting political orientation and
  measuring political homophily in twitter using big data.
\newblock {\em Journal of communication}, 64(2):317--332, 2014.

\bibitem{conover2011predicting}
M.~D. Conover, B.~Gon{\c{c}}alves, J.~Ratkiewicz, A.~Flammini, and F.~Menczer.
\newblock Predicting the political alignment of twitter users.
\newblock In {\em IEEE Third International Conference on Privacy, Security,
  Risk and Trust (PASSAT) and IEEE Third International Conference on Social
  Computing (SocialCom)}, pages 192--199. IEEE, 2011.

\bibitem{DaiICCV17}
B.~Dai, S.~Fidler, R.~Urtasun, and D.~Lin.
\newblock Towards diverse and natural image descriptions via a conditional gan.
\newblock In {\em Proceedings of the IEEE International Conference on Computer
  Vision (ICCV)}, 2017.

\bibitem{doersch2012makes}
C.~Doersch, S.~Singh, A.~Gupta, J.~Sivic, and A.~Efros.
\newblock What makes paris look like paris?
\newblock {\em ACM Transactions on Graphics}, 31(4), 2012.

\bibitem{Donahue_2015_CVPR}
J.~Donahue, L.~Anne~Hendricks, S.~Guadarrama, M.~Rohrbach, S.~Venugopalan,
  K.~Saenko, and T.~Darrell.
\newblock Long-term recurrent convolutional networks for visual recognition and
  description.
\newblock In {\em Proceedings of the IEEE Conference on Computer Vision and
  Pattern Recognition (CVPR)}, June 2015.

\bibitem{atlantic}
T.~B. Edsall.
\newblock Studies: Conservatives are from mars, liberals are from venus,
  February 2012.
\newblock
  \url{https://www.theatlantic.com/politics/archive/2012/02/studies-conservatives-are-from-mars-liberals-are-from-venus/252416/}.

\bibitem{Eisenschtat_2017_CVPR}
A.~Eisenschtat and L.~Wolf.
\newblock Linking image and text with 2-way nets.
\newblock In {\em Proceedings of the IEEE Conference on Computer Vision and
  Pattern Recognition (CVPR)}, 2017.

\bibitem{elliott2017imagination}
D.~Elliott and {\'A}.~K{\'a}d{\'a}r.
\newblock Imagination improves multimodal translation.
\newblock In {\em Proceedings of the Eighth International Joint Conference on
  Natural Language Processing (Volume 1: Long Papers)}, pages 130--141, 2017.

\bibitem{fei2005bayesian}
L.~Fei-Fei and P.~Perona.
\newblock A bayesian hierarchical model for learning natural scene categories.
\newblock In {\em Proceedings of the IEEE Conference on Computer Vision and
  Pattern Recognition (CVPR)}, volume~2, pages 524--531. IEEE, 2005.

\bibitem{symspell}
W.~Garbe.
\newblock Symspell.
\newblock \url{https://github.com/wolfgarbe/SymSpell}.

\bibitem{gilens1996race}
M.~Gilens.
\newblock Race and poverty in americapublic misperceptions and the american
  news media.
\newblock {\em Public Opinion Quarterly}, 60(4):515--541, 1996.

\bibitem{glorot2010understanding}
X.~Glorot and Y.~Bengio.
\newblock Understanding the difficulty of training deep feedforward neural
  networks.
\newblock In {\em Proceedings of the Thirteenth International Conference on
  Artificial Intelligence and Statistics (AISTATS)}, pages 249--256, 2010.

\bibitem{Gomez_2017_CVPR}
L.~Gomez, Y.~Patel, M.~Rusinol, D.~Karatzas, and C.~V. Jawahar.
\newblock Self-supervised learning of visual features through embedding images
  into text topic spaces.
\newblock In {\em Proceedings of the IEEE Conference on Computer Vision and
  Pattern Recognition (CVPR)}, 2017.

\bibitem{happer2013role}
C.~Happer and G.~Philo.
\newblock The role of the media in the construction of public belief and social
  change.
\newblock {\em Journal of Social and Political Psychology}, 1(1):321--336,
  2013.

\bibitem{he2016deep}
K.~He, X.~Zhang, S.~Ren, and J.~Sun.
\newblock Deep residual learning for image recognition.
\newblock In {\em Proceedings of the IEEE Conference on Computer Vision and
  Pattern Recognition (CVPR)}, pages 770--778, 2016.

\bibitem{hoffman2016learning}
J.~Hoffman, S.~Gupta, and T.~Darrell.
\newblock Learning with side information through modality hallucination.
\newblock In {\em Proceedings of the IEEE Conference on Computer Vision and
  Pattern Recognition (CVPR)}, pages 826--834. IEEE, 2016.

\bibitem{Hussain_2017_CVPR}
Z.~Hussain, M.~Zhang, X.~Zhang, K.~Ye, C.~Thomas, Z.~Agha, N.~Ong, and
  A.~Kovashka.
\newblock Automatic understanding of image and video advertisements.
\newblock In {\em Proceedings of the IEEE Conference on Computer Vision and
  Pattern Recognition (CVPR)}, July 2017.

\bibitem{jae2013style}
Y.~Jae~Lee, A.~A. Efros, and M.~Hebert.
\newblock Style-aware mid-level representation for discovering visual
  connections in space and time.
\newblock In {\em Proceedings of the IEEE International Conference on Computer
  Vision (ICCV)}, pages 1857--1864, 2013.

\bibitem{jiang2015self}
L.~Jiang, D.~Meng, Q.~Zhao, S.~Shan, and A.~G. Hauptmann.
\newblock Self-paced curriculum learning.
\newblock In {\em Twenty-Ninth Association for the Advancement of Artificial
  Intelligence (AAAI) Conference on Artificial Intelligence}, volume~2, page~6,
  2015.

\bibitem{jiang2018mentornet}
L.~Jiang, Z.~Zhou, T.~Leung, L.-J. Li, and L.~Fei-Fei.
\newblock Mentornet: Learning data-driven curriculum for very deep neural
  networks on corrupted labels.
\newblock In {\em Proceedings of the International Conference on Machine
  Learning (ICML)}, pages 2309--2318, 2018.

\bibitem{Johnson_2016_CVPR}
J.~Johnson, A.~Karpathy, and L.~Fei-Fei.
\newblock Densecap: Fully convolutional localization networks for dense
  captioning.
\newblock In {\em Proceedings of the IEEE Conference on Computer Vision and
  Pattern Recognition (CVPR)}, June 2016.

\bibitem{joo2014visual}
J.~Joo, W.~Li, F.~F. Steen, and S.-C. Zhu.
\newblock Visual persuasion: Inferring communicative intents of images.
\newblock In {\em Proceedings of the IEEE Conference on Computer Vision and
  Pattern Recognition (CVPR)}, 2014.

\bibitem{joo2015automated}
J.~Joo, F.~F. Steen, and S.-C. Zhu.
\newblock Automated facial trait judgment and election outcome prediction:
  Social dimensions of face.
\newblock In {\em Proceedings of the IEEE International Conference on Computer
  Vision (ICCV)}, 2015.

\bibitem{dlib09}
D.~E. King.
\newblock Dlib-ml: A machine learning toolkit.
\newblock {\em Journal of Machine Learning Research}, 10:1755--1758, 2009.

\bibitem{kingma2014adam}
D.~P. Kingma and J.~Ba.
\newblock Adam: A method for stochastic optimization.
\newblock {\em Proceedings of the International Conference on Learning
  Representations (ICLR)}, 2015.

\bibitem{Lambert_2018_CVPR}
J.~Lambert, O.~Sener, and S.~Savarese.
\newblock Deep learning under privileged information using heteroscedastic
  dropout.
\newblock In {\em Proceedings of the IEEE Conference on Computer Vision and
  Pattern Recognition (CVPR)}, June 2018.

\bibitem{le2014distributed}
Q.~Le and T.~Mikolov.
\newblock Distributed representations of sentences and documents.
\newblock In {\em Proceedings of the International Conference on Machine
  Learning (ICML)}, pages 1188--1196, 2014.

\bibitem{li2018patternnet}
H.~Li, J.~G. Ellis, L.~Zhang, and S.-F. Chang.
\newblock Patternnet: Visual pattern mining with deep neural network.
\newblock In {\em Proceedings of the 2018 ACM on International Conference on
  Multimedia Retrieval}, pages 291--299. ACM, 2018.

\bibitem{li2017mining}
Y.~Li, L.~Liu, C.~Shen, and A.~Van Den~Hengel.
\newblock Mining mid-level visual patterns with deep cnn activations.
\newblock {\em International Journal of Computer Vision (IJCV)},
  121(3):344--364, 2017.

\bibitem{malkov2016efficient}
Y.~A. Malkov and D.~A. Yashunin.
\newblock Efficient and robust approximate nearest neighbor search using
  hierarchical navigable small world graphs.
\newblock {\em IEEE Transactions on Pattern Analysis and Machine Intelligence
  (PAMI)}, 2016.

\bibitem{merrill_2016}
J.~B. Merrill.
\newblock Liberal, moderate or conservative? see how facebook labels you.
\newblock {\em The New York Times}, Aug 2016.

\bibitem{Liao2018Text}
B.~S. Minghui~Liao and X.~Bai.
\newblock {TextBoxes++}: A single-shot oriented scene text detector.
\newblock {\em {IEEE} Transactions on Image Processing}, 27(8):3676--3690,
  2018.

\bibitem{morin2005hierarchical}
F.~Morin and Y.~Bengio.
\newblock Hierarchical probabilistic neural network language model.
\newblock In {\em Tenth International Workshop on Artificial Intelligence and
  Statistics (AISTATS)}, volume~5, pages 246--252. Citeseer, 2005.

\bibitem{motiian2016information}
S.~Motiian, M.~Piccirilli, D.~A. Adjeroh, and G.~Doretto.
\newblock Information bottleneck learning using privileged information for
  visual recognition.
\newblock In {\em Proceedings of the IEEE Conference on Computer Vision and
  Pattern Recognition (CVPR)}, pages 1496--1505. IEEE, 2016.

\bibitem{munoz2017image}
C.~L. Mu{\~n}oz and T.~L. Towner.
\newblock The image is the message: Instagram marketing and the 2016
  presidential primary season.
\newblock {\em Journal of Political Marketing}, 16(3-4):290--318, 2017.

\bibitem{oquab2015object}
M.~Oquab, L.~Bottou, I.~Laptev, and J.~Sivic.
\newblock Is object localization for free?-weakly-supervised learning with
  convolutional neural networks.
\newblock In {\em Proceedings of the IEEE Conference on Computer Vision and
  Pattern Recognition (CVPR)}, pages 685--694, 2015.

\bibitem{paszke2017automatic}
A.~Paszke, S.~Gross, S.~Chintala, G.~Chanan, E.~Yang, Z.~DeVito, Z.~Lin,
  A.~Desmaison, L.~Antiga, and A.~Lerer.
\newblock Automatic differentiation in pytorch.
\newblock In {\em Advances in Neural Information Processing Systems Workshops
  (NIPS-W)}, 2017.

\bibitem{isidewith}
T.~Peck and N.~Boutelier.
\newblock Big political data.
\newblock \url{https://www.isidewith.com/polls}.
\newblock Accessed 2018.

\bibitem{Pedersoli_2017_ICCV}
M.~Pedersoli, T.~Lucas, C.~Schmid, and J.~Verbeek.
\newblock Areas of attention for image captioning.
\newblock In {\em Proceedings of the IEEE International Conference on Computer
  Vision (ICCV)}, Oct 2017.

\bibitem{peng2018same}
Y.~Peng.
\newblock Same candidates, different faces: Uncovering media bias in visual
  portrayals of presidential candidates with computer vision.
\newblock {\em Journal of Communication}, 68(5):920--941, 2018.

\bibitem{pennacchiotti2011machine}
M.~Pennacchiotti and A.-M. Popescu.
\newblock A machine learning approach to twitter user classification.
\newblock In {\em Fifth International Association for the Advancement of
  Artificial Intelligence (AAAI) Conference on Weblogs and Social Media}, 2011.

\bibitem{pentina2015curriculum}
A.~Pentina, V.~Sharmanska, and C.~H. Lampert.
\newblock Curriculum learning of multiple tasks.
\newblock In {\em Proceedings of the IEEE Conference on Computer Vision and
  Pattern Recognition (CVPR)}, pages 5492--5500, 2015.

\bibitem{peters2013content}
M.~E. Peters and D.~Lecocq.
\newblock Content extraction using diverse feature sets.
\newblock In {\em Proceedings of the 22nd International Conference on World
  Wide Web (WWW)}, pages 89--90. ACM, 2013.

\bibitem{philo2008active}
G.~Philo.
\newblock Active audiences and the construction of public knowledge.
\newblock {\em Journalism Studies}, 9(4):535--544, 2008.

\bibitem{recasens2013linguistic}
M.~Recasens, C.~Danescu-Niculescu-Mizil, and D.~Jurafsky.
\newblock Linguistic models for analyzing and detecting biased language.
\newblock In {\em Proceedings of the 51st Annual Meeting of the Association for
  Computational Linguistics (Volume 1: Long Papers)}, volume~1, pages
  1650--1659, 2013.

\bibitem{rehurek_lrec}
R.~{\v R}eh{\r u}{\v r}ek and P.~Sojka.
\newblock {Software Framework for Topic Modelling with Large Corpora}.
\newblock In {\em {Proceedings of the LREC 2010 Workshop on New Challenges for
  NLP Frameworks}}, pages 45--50, Valletta, Malta, May 2010. ELRA.
\newblock \url{http://is.muni.cz/publication/884893/en}.

\bibitem{richard2017weakly}
A.~Richard, H.~Kuehne, and J.~Gall.
\newblock Weakly supervised action learning with rnn based fine-to-coarse
  modeling.
\newblock In {\em Proceedings of the IEEE Conference on Computer Vision and
  Pattern Recognition (CVPR)}, pages 754--763, 2017.

\bibitem{schill2012visual}
D.~Schill.
\newblock The visual image and the political image: A review of visual
  communication research in the field of political communication.
\newblock {\em Review of Communication}, 12(2):118--142, 2012.

\bibitem{schreiber2013red}
D.~Schreiber, G.~Fonzo, A.~N. Simmons, C.~T. Dawes, T.~Flagan, J.~H. Fowler,
  and M.~P. Paulus.
\newblock Red brain, blue brain: Evaluative processes differ in democrats and
  republicans.
\newblock {\em PLOS ONE}, 8(2):1--6, 02 2013.

\bibitem{schroff2015facenet}
F.~Schroff, D.~Kalenichenko, and J.~Philbin.
\newblock Facenet: A unified embedding for face recognition and clustering.
\newblock In {\em Proceedings of the IEEE Conference on Computer Vision and
  Pattern Recognition (CVPR)}, pages 815--823, 2015.

\bibitem{sharmanska2013learning}
V.~Sharmanska, N.~Quadrianto, and C.~H. Lampert.
\newblock Learning to rank using privileged information.
\newblock In {\em Proceedings of the IEEE Conference on Computer Vision and
  Pattern Recognition (CVPR)}, pages 825--832. IEEE, 2013.

\bibitem{sicre2017unsupervised}
R.~Sicre, Y.~S. Avrithis, E.~Kijak, and F.~Jurie.
\newblock Unsupervised part learning for visual recognition.
\newblock In {\em Proceedings of the IEEE Conference on Computer Vision and
  Pattern Recognition (CVPR)}, pages 3116--3124, 2017.

\bibitem{singh2012unsupervised}
S.~Singh, A.~Gupta, and A.~A. Efros.
\newblock Unsupervised discovery of mid-level discriminative patches.
\newblock In {\em Proceedings of the European Conference on Computer Vision
  (ECCV)}, pages 73--86. Springer, 2012.

\bibitem{sivic2005discovering}
J.~Sivic, B.~C. Russell, A.~A. Efros, A.~Zisserman, and W.~T. Freeman.
\newblock Discovering objects and their location in images.
\newblock In {\em Proceedings of the IEEE International Conference on Computer
  Vision (ICCV)}, volume~1, pages 370--377. IEEE, 2005.

\bibitem{chris_bmvc2018}
C.~Thomas and A.~Kovashka.
\newblock Persuasive faces: Generating faces in advertisements.
\newblock In {\em Proceedings of the British Machine Vision Conference (BMVC)},
  2018.

\bibitem{vapnik2015learning}
V.~Vapnik and R.~Izmailov.
\newblock Learning using privileged information: similarity control and
  knowledge transfer.
\newblock {\em Journal of Machine Learning Research (JMLR)}, 16(2023-2049):2,
  2015.

\bibitem{Venugopalan_2017_CVPR}
S.~Venugopalan, L.~Anne~Hendricks, M.~Rohrbach, R.~Mooney, T.~Darrell, and
  K.~Saenko.
\newblock Captioning images with diverse objects.
\newblock In {\em Proceedings of the IEEE Conference on Computer Vision and
  Pattern Recognition (CVPR)}, July 2017.

\bibitem{vinyals2015show}
O.~Vinyals, A.~Toshev, S.~Bengio, and D.~Erhan.
\newblock Show and tell: A neural image caption generator.
\newblock In {\em Proceedings of the IEEE Conference on Computer Vision and
  Pattern Recognition (CVPR)}, pages 3156--3164, 2015.

\bibitem{volkova2014inferring}
S.~Volkova, G.~Coppersmith, and B.~Van~Durme.
\newblock Inferring user political preferences from streaming communications.
\newblock In {\em Proceedings of the 52nd Annual Meeting of the Association for
  Computational Linguistics (Volume 1: Long Papers)}, volume~1, pages 186--196,
  2014.

\bibitem{wang2017untrimmednets}
L.~Wang, Y.~Xiong, D.~Lin, and L.~Van~Gool.
\newblock Untrimmednets for weakly supervised action recognition and detection.
\newblock In {\em Proceedings of the IEEE Conference on Computer Vision and
  Pattern Recognition (CVPR)}, pages 4325--4334, 2017.

\bibitem{wang2017polarized}
Y.~Wang, Y.~Feng, Z.~Hong, R.~Berger, and J.~Luo.
\newblock How polarized have we become? a multimodal classification of trump
  followers and clinton followers.
\newblock In {\em International Conference on Social Informatics}, 2017.

\bibitem{wang2016deciphering}
Y.~Wang, Y.~Li, and J.~Luo.
\newblock Deciphering the 2016 us presidential campaign in the twitter sphere:
  A comparison of the trumpists and clintonists.
\newblock In {\em Tenth International Association for the Advancement of
  Artificial Intelligence (AAAI) Conference on Web and Social Media}, pages
  723--726, 2016.

\bibitem{wei2018ts2c}
Y.~Wei, Z.~Shen, B.~Cheng, H.~Shi, J.~Xiong, J.~Feng, and T.~Huang.
\newblock Ts2c: Tight box mining with surrounding segmentation context for
  weakly supervised object detection.
\newblock In {\em Proceedings of the European Conference on Computer Vision
  (ECCV)}, pages 434--450, 2018.

\bibitem{wong2016quantifying}
F.~M.~F. Wong, C.~W. Tan, S.~Sen, and M.~Chiang.
\newblock Quantifying political leaning from tweets, retweets, and retweeters.
\newblock {\em IEEE Transactions on Knowledge and Data Engineering},
  28(8):2158--2172, 2016.

\bibitem{yepami2019}
K.~{Ye}, N.~{Honarvar Nazari}, J.~{Hahn}, Z.~{Hussain}, M.~{Zhang}, and
  A.~{Kovashka}.
\newblock Interpreting the rhetoric of visual advertisements.
\newblock {\em To appear, IEEE Transactions on Pattern Analysis and Machine
  Intelligence (PAMI)}, 2019.

\bibitem{Ye_2019_ICCV}
K.~Ye, M.~Zhang, A.~Kovashka, W.~Li, D.~Qin, and J.~Berent.
\newblock Cap2det: Learning to amplify weak caption supervision for object
  detection.
\newblock In {\em Proceedings of the IEEE International Conference on Computer
  Vision (ICCV)}, Oct 2019.

\bibitem{zamir2017feedback}
A.~R. Zamir, T.-L. Wu, L.~Sun, W.~B. Shen, B.~E. Shi, J.~Malik, and
  S.~Savarese.
\newblock Feedback networks.
\newblock In {\em Proceedings of the IEEE Conference on Computer Vision and
  Pattern Recognition (CVPR)}, pages 1808--1817. IEEE, 2017.

\bibitem{zhang2017curriculum}
Y.~Zhang, P.~David, and B.~Gong.
\newblock Curriculum domain adaptation for semantic segmentation of urban
  scenes.
\newblock In {\em Proceedings of the IEEE International Conference on Computer
  Vision (ICCV)}, pages 2020--2030, 2017.

\bibitem{zhou2016learning}
B.~Zhou, A.~Khosla, A.~Lapedriza, A.~Oliva, and A.~Torralba.
\newblock Learning deep features for discriminative localization.
\newblock In {\em Proceedings of the IEEE Conference on Computer Vision and
  Pattern Recognition (CVPR)}, pages 2921--2929, 2016.

\bibitem{zhou2010unsupervised}
F.~Zhou, F.~De~la Torre, and J.~F. Cohn.
\newblock Unsupervised discovery of facial events.
\newblock In {\em Proceedings of the IEEE Conference on Computer Vision and
  Pattern Recognition (CVPR)}, pages 2574--2581. IEEE, 2010.

\end{thebibliography}


\begin{thebibliography}{10}\itemsep=-1pt

\bibitem{le2014distributed_1}
Q.~Le and T.~Mikolov.
\newblock Distributed representations of sentences and documents.
\newblock In {\em Proceedings of the International Conference on Machine
  Learning (ICML)}, pages 1188--1196, 2014.

\end{thebibliography}
}

\end{document}


\maketitle

\section{Introduction}
This document presents supplementary results to our main text. We first present additional details of our new political bias dataset, in Section \ref{sec:dataset}. Next, in Section \ref{sec:quantitative}, we provide two additional quantitative results using our test set which shows the differences between our best performing method and the baselines on the various topics within our dataset. We also provide results for an application of predicting the bias of different media sources. In Section \ref{sec:img_to_word}, we present additional qualitative results to complement our result in Fig.~5 from the main text, i.e.~images that most strongly predicted several words from articles. In Section \ref{sec:doc2vec}, we illustrate what our trained document embedding model learns by showing nearby words for a number of query words. In Section \ref{sec:gng}, we compare human vs.~machine performance by showing images that either our best algorithm or humans failed to classify (or both). In Section \ref{sec:mturk}, we include additional  examples of images agreed upon by human annotators, as well as the free-form text reasons our participants gave for their Left / Right guesses. We also include our MTurk data collection interface. 
Finally, in Section \ref{sec:dataset_examples}, we show example images and articles from our dataset.

\section{Dataset Details}
\label{sec:dataset}
In this section, we present additional details of our new political bias dataset to complement our main text. Our dataset contains 1,861,336 images total and 1,559,004 articles total. However, after our deduplication procedure (described in our main text), we are left with 1,079,588 unique images upon which we conduct all experiments. In this section, we break down this \textit{unique} count by politics, topic, and media source. We wish to re-emphasize that even though we exclude duplicates here, the articles associated with duplicate images are not necessarily duplicates (the overwhelming majority are unique). Thus, a large body of potentially useful image-text pairs are excluded from this description because the image associated with the text is not unique. 

\begin{figure}[t]
    \centering
    \includegraphics[height=2in]{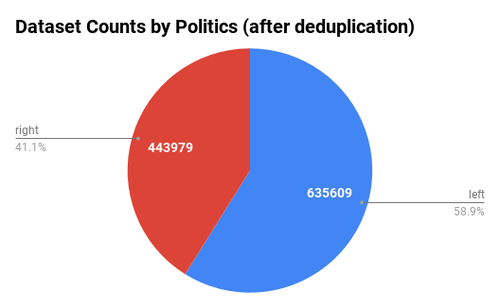}
    \caption{We illustrate the distribution of Left/Right unique images in our deduplicated dataset.}
    \label{fig:politics_counts}
\end{figure}

\begin{figure}[t]
    \centering
    \includegraphics[width=1\textwidth]{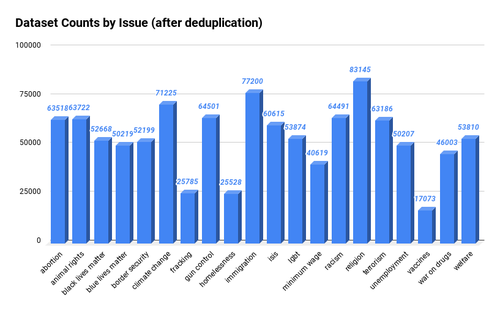}
    \caption{We show the distribution of unique images in our dataset by topic, across both Left/Right.}
    \label{fig:topic_counts}
\end{figure}

Figure \ref{fig:politics_counts} shows the breakdown of unique images in our dataset by politics. There are more images on the left than on the right, resulting in a slight class imbalance. We correct for this class imbalance during training for all of our experiments by ensuring equal class weight in the loss terms. Figure \ref{fig:topic_counts} further breaks down the distribution images by topic. For example, we see our dataset contains 83,145 unique images on the subject of religion (from both L/R), our most frequent category, while we collected 17,073 on the subject of vaccines, our least frequent category.

We also present the frequency distribution of our deduplicated dataset broken down by media source in the attached Microsoft Excel file \textbf{\texttt{media\_source\_stats.xlsx}} as there are too many to include or visualize in this document. Note that we also include the political leaning of the media source, as assigned by Media Bias Fact Check (see our main text for details).

\section{Quantitative Results}
\label{sec:quantitative}
\begin{table*}
    \centering
    \resizebox{1\textwidth}{!}{
    \begin{tabular}{|c||c|c|c||c|c|c|}
    \hline
    Method & Vaccines & Fracking & War on Drugs & Border Security & Black Lives Matter & Climate Change \\
    \hline
    \textsc{Resnet} & 0.6768 & 0.6737 & 0.6684 & 0.6922 & 0.7026 & 0.6934 \\
    \hline
    \textsc{Ours} & 0.7422 & 0.7209 & 0.7128 & 0.7161 & 0.7269 & 0.7179 \\    
    \hline
    \end{tabular}
    }
    \caption{Average performance for the three topics where our method achieves the largest vs smallest improvement over the baseline.}
    \label{tab:topic}
\end{table*}
We present two quantitative results to supplement our main text. We first wanted to understand on what types of images our best performing method, \textsc{Ours} outperformed the \textsc{Resnet} baseline. In Table \ref{tab:topic}, we show a result which shows the top-3 topics that our method performed the best (and worst) over the baseline. We notice that for no topic does the baseline outperform our method. Even for those topics on which the baseline performs most competitively with our method, our method still outperforms it by 1-2\%.  We include complete results including additional baselines, for all topics in the included file, \textbf{\texttt{topic\_results.xlsx}}. 

\begin{table*}[t]
    \centering
    \resizebox{1\textwidth}{!}{
    \begin{tabular}{|c||c|c||c|c|c|c|c|c|c|c|c|}
    \hline
    Method & Top-20 & Top-100 & Sun & Change & Breitbart & NewSt & NewYorker & NatRev & Slate & CNN & RevCom \\
    \hline
    \textsc{Resnet} & 0.697 & 0.690  & 0.627 & 0.653 & 0.527 & \textbf{0.821} & 0.873 & 0.718 & \textbf{0.798} & 0.795 & \textbf{0.875}\\
    \hline
    \textsc{Ours} & \textbf{0.739} & \textbf{0.724} & \textbf{0.707} & \textbf{0.690} & \textbf{0.607} & 0.808 & \textbf{0.934} & \textbf{0.758} & 0.793 & \textbf{0.873} & 0.781 \\
    \hline
    \end{tabular}
    }
    \caption{Average performance for the top-20, and top-100 news sources, and individual results for some popular news sources.}
    \label{tab:source}
\end{table*}

In Table \ref{tab:source}, we analyze the results as a function of the media source to which the image belongs. We compute the performance of our method on images exclusively from a particular media source, for each media source.
We then rank the sources by number of samples in the test set, and check how performance changes as the number of samples decreases. We see that for media sources with more samples, \textsc{Ours} achieves a stronger result than the \textsc{Resnet} baseline (0.739 vs 0.697). 
We also show results for individual well-known media sources that have many samples in our dataset. The Sun, Breitbart, and National Review are well-known right-leaning sources, while the rest are left-leaning. 
Our method works well for both right- and left-leaning sources. For a few left-leaning sources, the baseline achieves stronger results. Among common sources, the baseline's gain is largest on RevCom, a \emph{very} far-left, ``revolutionary communism'' website.
It is surprising to see how accurate we can infer leaning from images alone; close to or over 80\% for many sources shown. 

We also provide supplementary results to complement this result in \textbf{\texttt{media\_source\_results.xlsx}}, including for other baselines. We break down the performance for each of our methods by media source. We observe that our method, \textsc{Ours} consistently outperforms the baselines, often substantially. 
\section{Image to Word Prediction Results}
\label{sec:img_to_word}
In our main text, we described a model trained to predict words from images. We trained this model to predict which words, from a fixed dictionary of the 1000 most visual words (see main text for details), would be in the article paired with the image. For this result only, we also conditioned the model on the document embedding of the article paired with the image. After training, we ran our entire large weakly-supervised test set through this model and predicted words for all images. For each word, we then sorted all test set images by the score the model assigned for the prediction of that word and show the 100 images for each that have the highest overall probability. We include results for several words in the {\textbf{\texttt{image\_to\_word}}} folder. We include results for several words, including ``immigrant'', ``lgbt'', ``antifa'', and ``nationalist''. We see that the images which strongly predicted the word ``immigrant'' often feature Hispanic people, children, or law enforcement symbols / personnel. For ``lgbt'', we notice that many images feature rainbow flags. ``Antifa'' often features street scenes with protestors wearing black. We also observe fascist symbols, such as swastikas or Nazi salutes in these photos. ``Nationalist'' features numerous examples of white supremacist imagery, including Ku Klux Klan garbs, swastikas, and Celtic crosses: symbolism associated with white supremacist and neo-nazi movements. Collectively these results indicate that, although the articles paired with the text are lengthly and much more weakly aligned than traditional image-text embedding tasks (i.e.~captions, descriptions, etc.), a consistent visual signal exists that our model is able to grasp.
\section{Textual Embedding Word Retrieval Results}

\begin{table}[t]
\resizebox{1\textwidth}{!}{
\begin{tabular}{|c|c|c|c|c|c|}
\hline
\textbf{Query phrase:} & donald trump & charlottesville & liberal & fascist & parkland \\ \hline
Results: & \begin{tabular}[c]{@{}c@{}}auxiliar: 0.4155\\ intensa: 0.4132\\ macron: 0.4102\\ putin: 0.4042\\ ultraderecha: 0.4010\\ horripilantes: 0.4005\\ billionaire: 0.3991\\ pence: 0.3980\\ obama: 0.3937\\ cruz: 0.3928\\ duterte: 0.3924\\ erdogan: 0.3919\\ continuado: 0.3898\\ mueller: 0.3876\\ tonight: 0.3874\\ inauguration: 0.3869\\ gop: 0.3852\\ america: 0.3848\\ potus: 0.3835\\ brexit: 0.3834\\ presidency: 0.3819\\ alabama: 0.3817\\ marcharse: 0.3814\\ cabinet: 0.3812\\ netanyahu: 0.3779\\ milo: 0.3770\\ republicans: 0.3766\\ opioid: 0.3757\\ comey: 0.3737\\ \#: 0.3736\end{tabular} & \begin{tabular}[c]{@{}c@{}}charleston: 0.7303\\ parkland: 0.7189\\ antifa: 0.7135\\ kkk: 0.7117\\ ferguson: 0.7038\\ dallas: 0.6998\\ confederate: 0.6995\\ richmond: 0.6956\\ shooting: 0.6879\\ horrific: 0.6844\\ portland: 0.6828\\ riots: 0.6826\\ cleveland: 0.6817\\ heyer: 0.6806\\ protest: 0.6782\\ rally: 0.6779\\ nfl: 0.6760\\ tragedy: 0.6757\\ islamophobia: 0.6727\\ anticom: 0.6721\\ spike: 0.6719\\ berkeley: 0.6718\\ counterprotesters: 0.6702\\ barcelona: 0.6692\\ memphis: 0.6679\\ heaphy: 0.6669\\ alt: 0.6665\\ weekend: 0.6662\\ mcauliffe: 0.6657\\ spencer: 0.6654\end{tabular} & \begin{tabular}[c]{@{}c@{}}leftist: 0.2721\\ progressive: 0.2650\\ conservative: 0.2583\\ liberals: 0.2541\\ much: 0.2516\\ wing: 0.2516\\ mainstream: 0.2514\\ centrist: 0.2420\\ moderate: 0.2323\\ emerged: 0.2312\\ dismal: 0.2309\\ steadily: 0.2269\\ radical: 0.2263\\ portrayed: 0.2256\\ conservatives: 0.2253\\ shifted: 0.2248\\ defeaning: 0.2245\\ plummeted: 0.2244\\ outflanked: 0.2219\\ progressives: 0.2218\\ leftwing: 0.2217\\ touted: 0.2209\\ democrat: 0.2209\\ 12,030: 0.2204\\ long: 0.2196\\ corporatists: 0.2194\\ served: 0.2186\\ framed: 0.2186\\ hardline: 0.2182\\ leftward: 0.2176\end{tabular} & \begin{tabular}[c]{@{}c@{}}fascism: 0.7861\\ fascists: 0.7494\\ nazi: 0.7169\\ racists: 0.7128\\ racist: 0.7068\\ totalitarian: 0.6903\\ repressive: 0.6866\\ terrorist: 0.6862\\ filmado: 0.6791\\ imperialist: 0.6771\\ communist: 0.6729\\ nazis: 0.6666\\ globalist: 0.6659\\ nationalist: 0.6655\\ genocidal: 0.6630\\ rogue: 0.6627\\ authoritarian: 0.6620\\ extremist: 0.6603\\ vanguard: 0.6599\\ antifascist: 0.6583\\ avakian: 0.6579\\ aholes: 0.6571\\ waok: 0.6566\\ troutdale: 0.6565\\ clown: 0.6564\\ supremacist: 0.6556\\ democrat: 0.6548\\ supremacy: 0.6548\\ lunatic: 0.6545\\ misogynist: 0.6533\end{tabular} & \begin{tabular}[c]{@{}c@{}}newtown: 0.7640\\ hogg: 0.7635\\ stoneman: 0.7501\\ nra: 0.7455\\ charlottesville: 0.7189\\ shooting: 0.7161\\ walkout: 0.7135\\ walkouts: 0.7029\\ charleston: 0.7002\\ tragedy: 0.6991\\ orlando: 0.6986\\ emma4change: 0.6931\\ msd: 0.6844\\ sandyhook: 0.6841\\ shootings: 0.6795\\ gun: 0.6752\\ marjory: 0.6739\\ senseless: 0.6701\\ kasky: 0.6688\\ neveragain: 0.6665\\ trayvon: 0.6654\\ 7to: 0.6644\\ sarasota: 0.6613\\ columbine: 0.6610\\ horrific: 0.6597\\ gaskill: 0.6596\\ manjarres: 0.6596\\ florida: 0.6583\\ loesch: 0.6576\\ nationalwalkoutday: 0.6574\end{tabular} \\ \hline
\end{tabular}
}
\caption{We show examples of our learned text embedding. At the top, we show several ``query phrases'' which we embed using our method. We then compute the distance from each query phrase to all other learned words in our dataset's vocabulary and rank the words in order of increasing distance. Thus, retrieved words near the top are more closely related to the query phrase in the learned space than words below.}
\label{tab:doc2vec}
\end{table}

\label{sec:doc2vec}
We trained a text embedding \cite{le2014distributed_1} model on articles from our dataset. In Table \ref{tab:doc2vec} we show an example of what our model learned for a number of query words. We compute the embedding of the query words using our model, then find the nearest words in embedding space from the learned dictionary and rank them. We observe that for ``Donald Trump'', several of the top words are in Spanish, which are likely coming from articles related to immigration concerning Trump. The translation of these words is fitting in this context, i.e. \textit{intensa} means ``intense'', while ``ultraderecha'' means far-right. ``Horripilantes'' means ''horrifying / terrifying.'' We also notice a ``\#'' sign associated with Trump, likely coming from his use of Twitter. Importantly, we noticed for \textit{events}, like Charlottesville (a protest event in which a protestor was run over by a car in a hate crime), relevant concepts that our \textit{image} classifiers could potentially pick up on appear. For example, ``riots'', ``antifa'' (a protest group), ``rally'', etc.\ are all visualizable concepts associated with the event. 
We observe for another event, ``Parkland'' (a mass school shooting event involving 17 deaths), nearby concepts are ``Newtown'' (another school shooting), ``Hogg'' (a survivor of the Parkland shooting), ``NRA'' (the National Rifle Association, which opposed gun measures following the event and was the subject of significant press), and a variety of other hashtags and concepts associated with the event. We believe that these results illustrate \textit{how} leveraging text helps our method ultimately perform better by forcing our classifiers to learn to pay attention to certain visual concepts, after being conditioned on the latent document embedding at training time.
The representation our image classifiers learn guided by this latent, weak supervision ultimately proves to be superior to the other approaches we tested. We include many additional word query results in \textbf{\texttt{learned\_word\_embeddings.xlsx}}.
\section{Human vs.\ Machine Abilities}
\label{sec:gng}
\begin{figure}[t]
    \centering
    \includegraphics[width=1\linewidth]{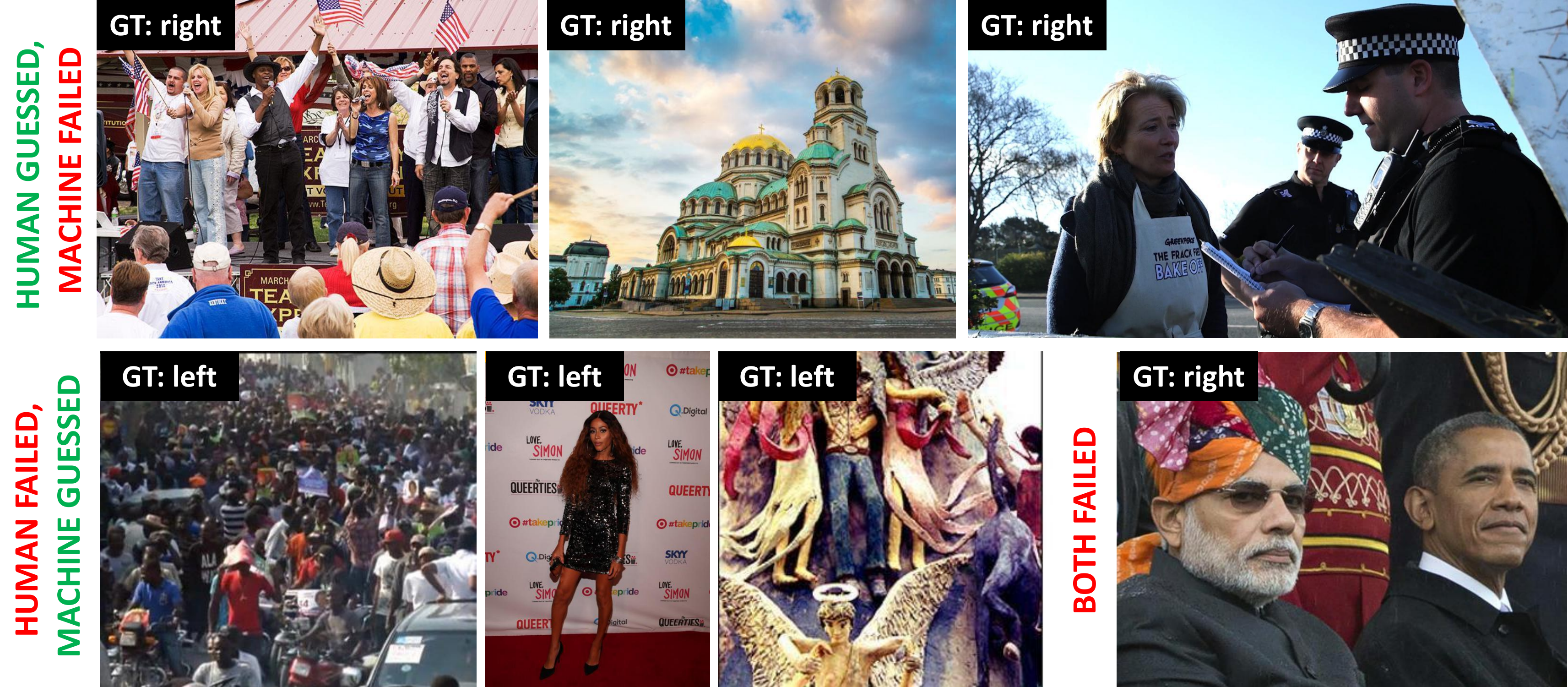}
    \caption{Images that either our best algorithm failed to classify (top), humans failed (bottom left) or both human and machine failed (bottom right). Please see text for our explanation.}
    \label{fig:gng}
    \vspace{-0.5em}
\end{figure}
In Fig.~\ref{fig:gng}, we show images that humans and/or our best-performing algorithm (\textsc{Ours}) were able/unable to classify. At the top, we show the gap between human reasoning abilities and our classifier. The first image at the top has a subtle country vibe (associated with the right), which was imperceptible for our algorithm. Next is an image of a non-western church, which was likely too different from churches in the training set. %
The third image shows British actress Emma Thompson campaigning for Greenpeace and getting cited; our algorithm is unable to analyze such complex scenes. 
At the bottom left of the figure, we show images that humans were unable to classify, but bias in the data helped our algorithm classify. Images of protests, Hollywood, and art, are common in left-leaning images. Finally, we show an image that neither human nor algorithm were able to classify, as it depends on context from the article, which is unavailable at test time. 

\section{MTurk Responses}
\label{sec:mturk}
\begin{figure}[!th]
    \centering
    \includegraphics[width=1\textwidth]{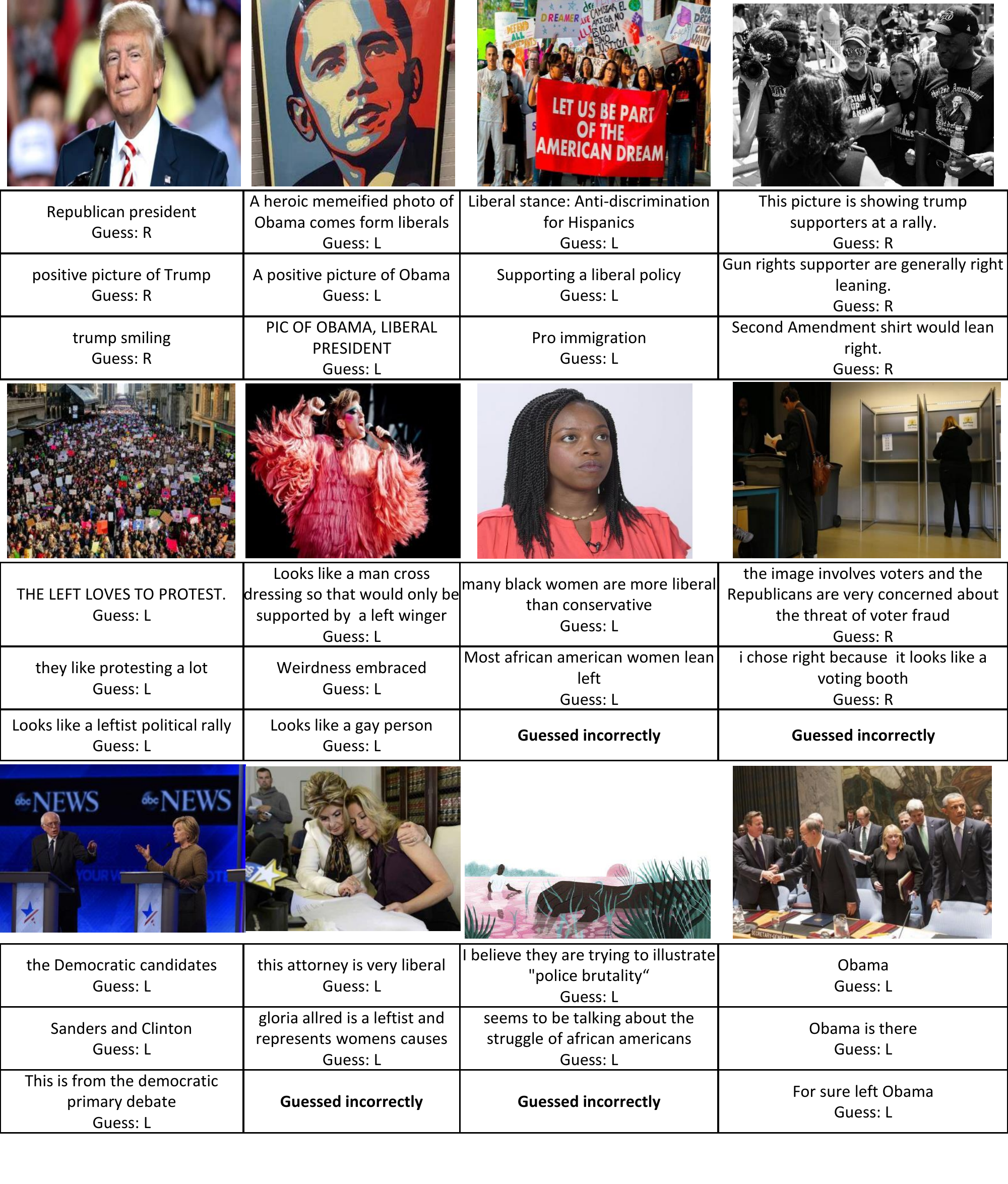}
    \caption{Examples of images whose politics were correctly guessed by at least a majority (2/3) of MTurkers. We also include the reasons given for their prediction by the MTurkers below each image. MTurkers who guessed the image incorrectly are indicated by ``Guessed incorrectly.''}
    \label{fig:mturK_ex1}
\end{figure}

\begin{figure}[!th]
    \centering
    \includegraphics[width=1\textwidth]{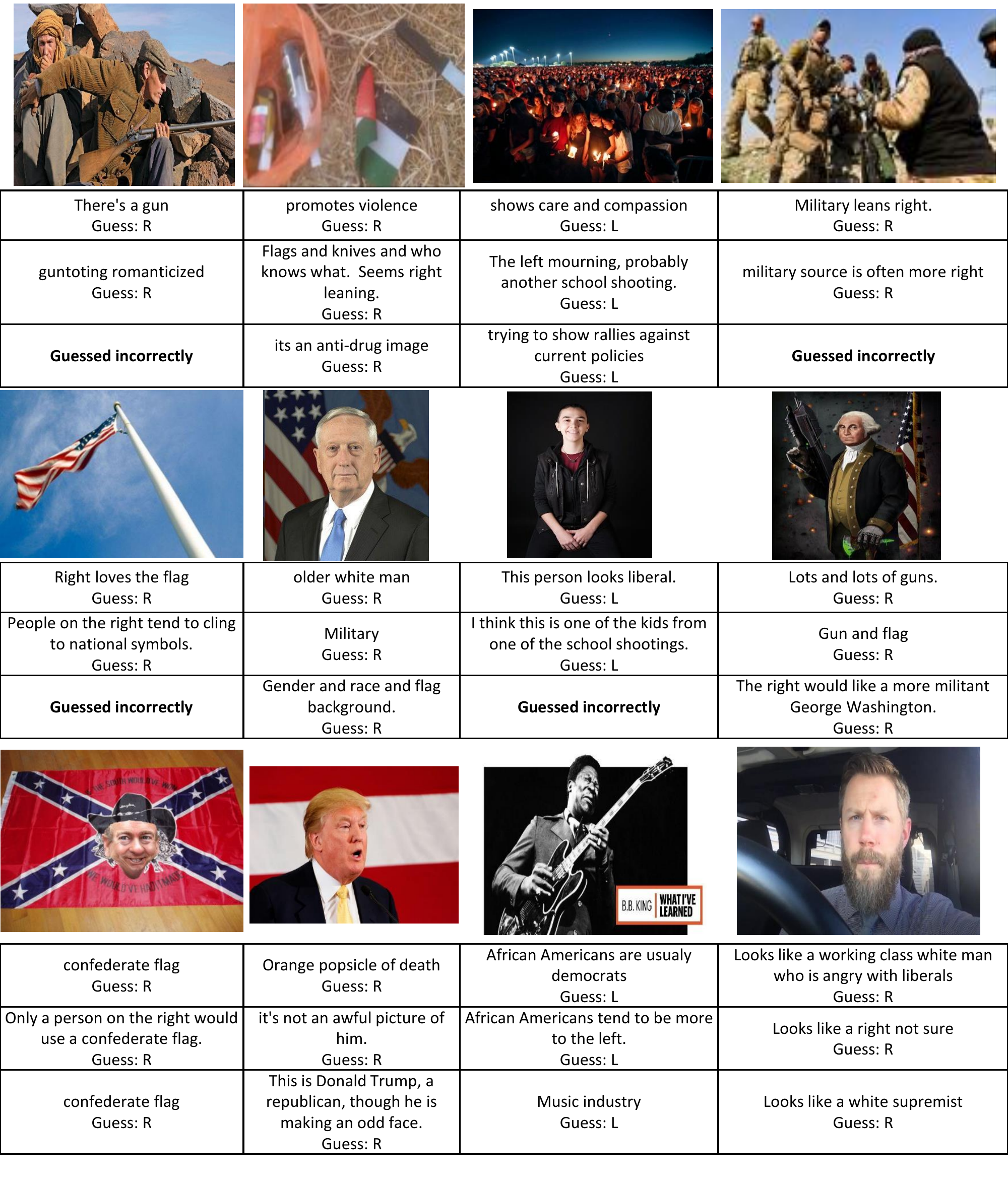}
    \caption{Examples of images whose politics were correctly guessed by at least a majority (2/3) of MTurkers. We also include the reasons given for their prediction by the MTurkers below each image. MTurkers who guessed the image incorrectly are indicated by ``Guessed incorrectly.''}
    \label{fig:mturK_ex2}
\end{figure}

\begin{figure}[!th]
    \centering
    \includegraphics[width=1\textwidth]{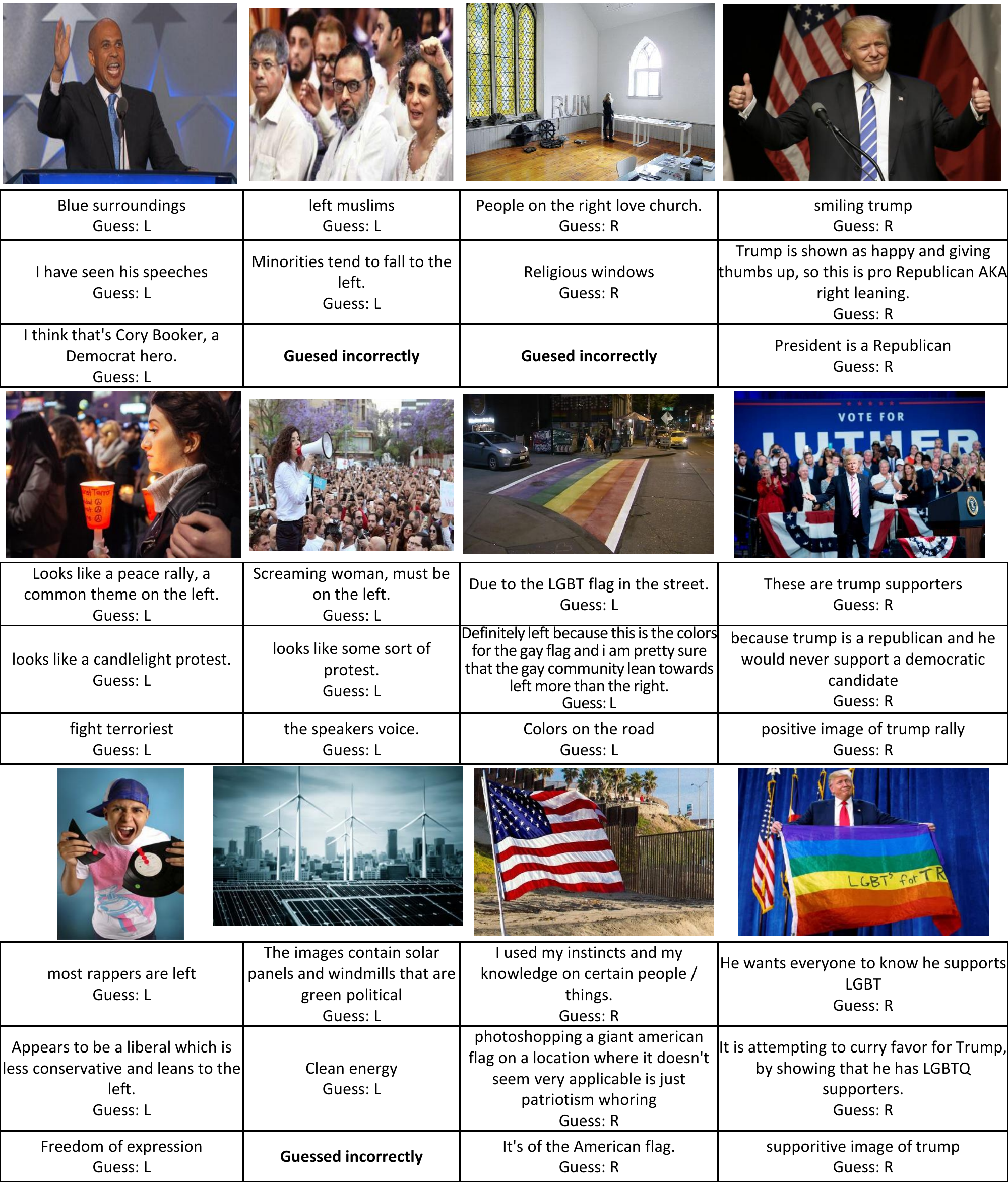}
    \caption{Examples of images whose politics were correctly guessed by at least a majority (2/3) of MTurkers. We also include the reasons given for their prediction by the MTurkers below each image. MTurkers who guessed the image incorrectly are indicated by ``Guessed incorrectly.''}
    \label{fig:mturK_ex3}
\end{figure}

In Figures \ref{fig:mturK_ex1}-\ref{fig:mturK_ex3}, we show example images which at least a majority (2/3) of humans were able to guess the politics of correctly. We note that many times, when a politician of a particular party is shown, human annotators assume the image is the same leaning as the politician's party (e.g.\ image of Trump is right). Annotators often rely on racial stereotypes as well (``black women are more liberal,'' ``most rappers are left,'' ``left muslims'', '``older white man'' is right-leaning). Relying on these stereotypical concepts in our \textsc{Human Concepts} model explains why that model performs best on those images containing humans (see main text), though it doesn't perform best in the dataset at large. We also observe that humans tend to associate the right with guns, patriotic symbols, and religion, whereas they tend to associate peace, compassion, diversity, protests, and minorities with the left. Humans also recognized some of the people appearing in the images and relied on their external knowledge of that person's political leanings to guess the image's label. We also include a complete listing of the concepts that were extracted from the MTurk free-form text explanations in \textbf{\texttt{human\_concepts.xlsx}}.

We also include our MTurk data collection interface in HTML file \textbf{\texttt{MTurk\_Inferface.html}}. Note that as you answer the questions, additional questions will appear. We first asked annotators to determine if the image met certain exclusionary criteria, i.e.~text, blurry, etc. We then asked annotators to classify the image as left / right / ambiguous. We then asked what features of the image were relied on by the annotator to make their decision. We then showed annotators the article text going with the document and asked whether the text met certain exclusionary criteria, mainly originating from HTML scraping errors. We also asked annotators if the image and text were related to one another and to paste the text from the article that most aligned with the image. We then asked the workers to predict the politics of the image-text \textit{pair} (as opposed to the image alone) and finally asked workers to state political topic(s) of the image-text pair.

\section{Example Images and Documents from Dataset}
\label{sec:dataset_examples}
\begin{figure}[!th]
    \centering
    \includegraphics[width=1\textwidth]{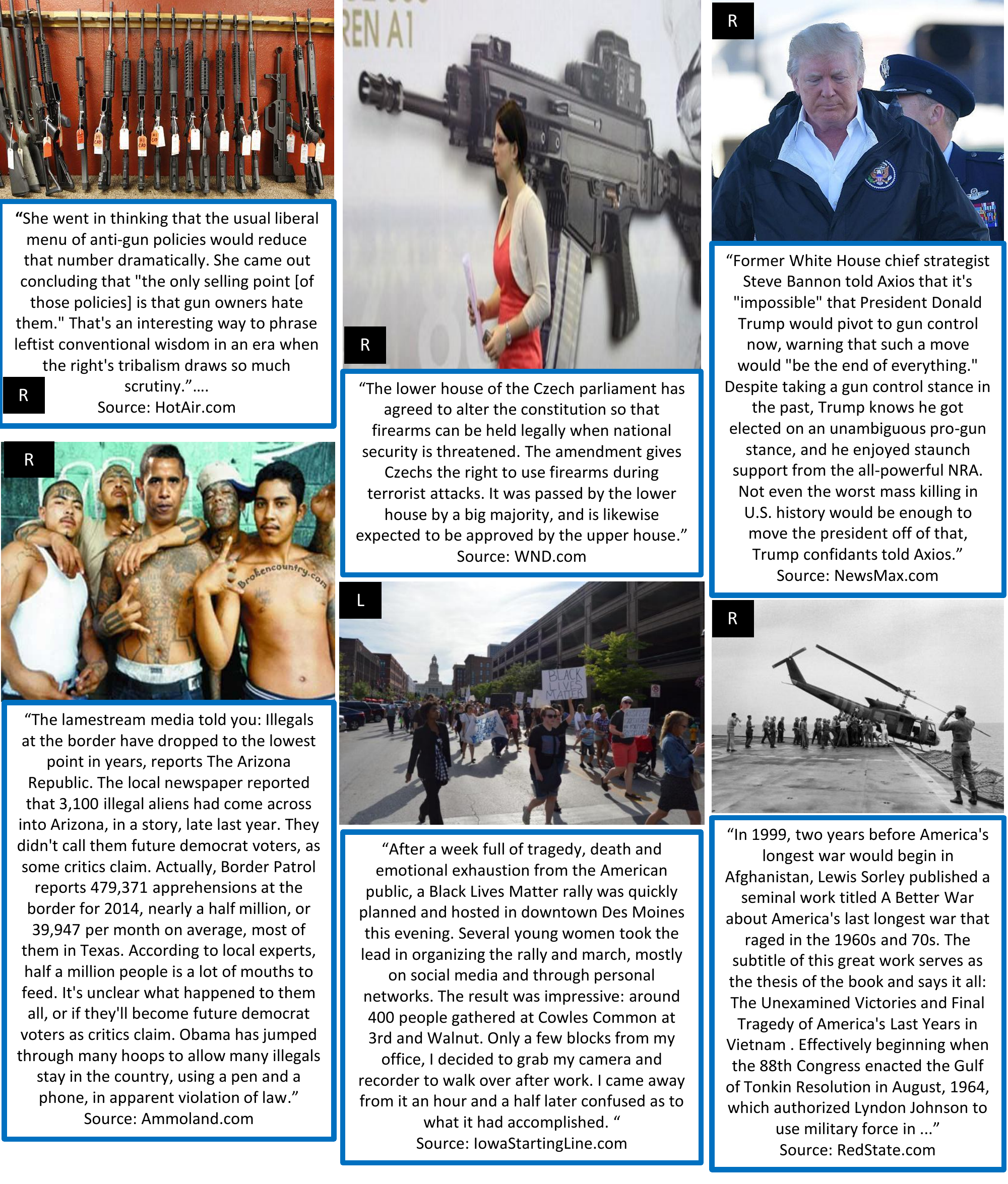}
    \caption{Example images and articles (truncated) from our dataset. We annotate each image with the media source from which it and the article came, as well as the politics of that media source, as determined by Media Bias Fact Check (see our main text for details).}
    \label{fig:dataset_ex1}
\end{figure}

\begin{figure}[!th]
    \centering
    \includegraphics[width=1\textwidth]{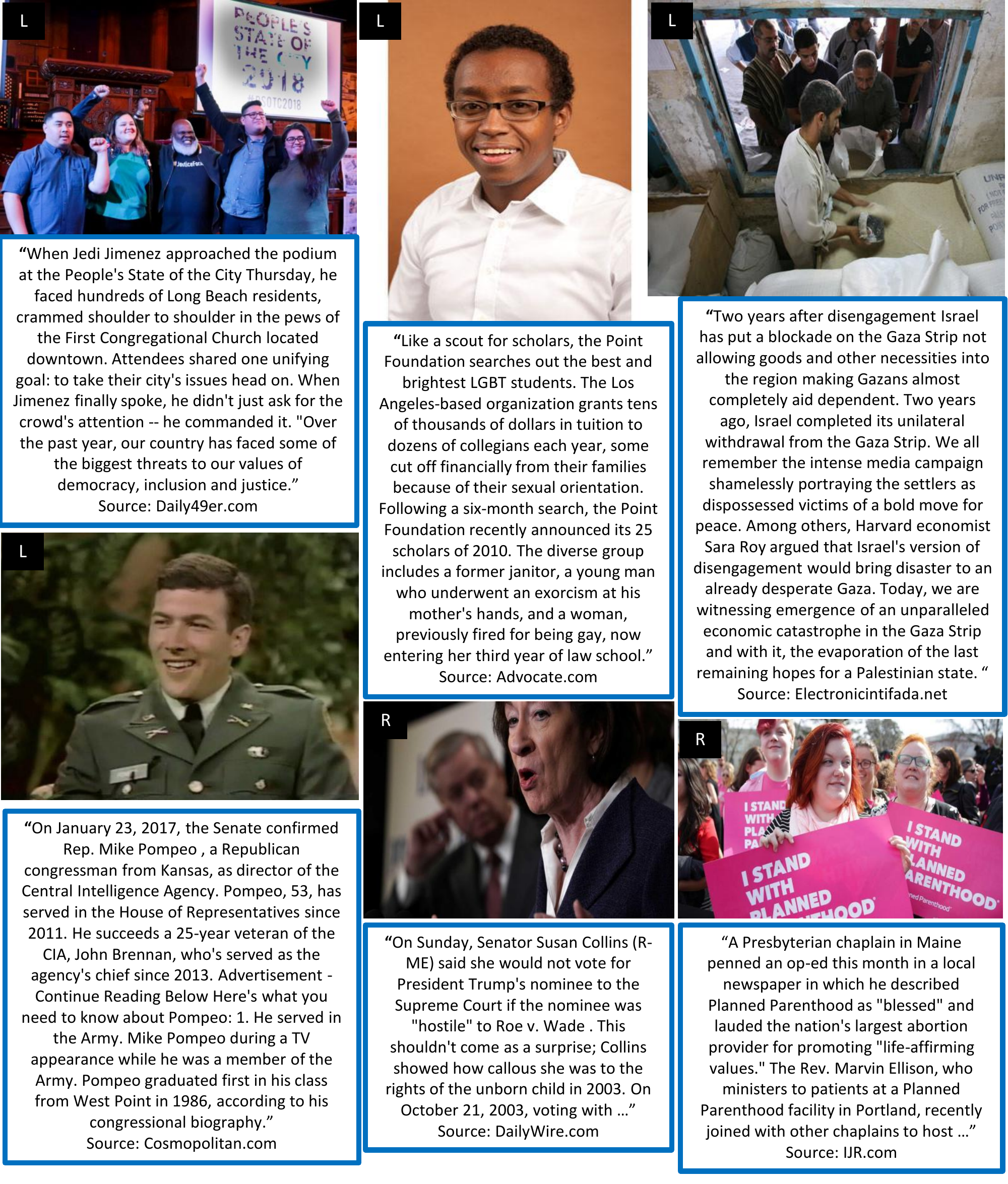}
    \caption{Example images and articles (truncated) from our dataset. We annotate each image with the media source from which it and the article came, as well as the politics of that media source, as determined by Media Bias Fact Check (see our main text for details).}
    \label{fig:dataset_ex2}
\end{figure}

\begin{figure}[!th]
    \centering
    \includegraphics[width=1\textwidth]{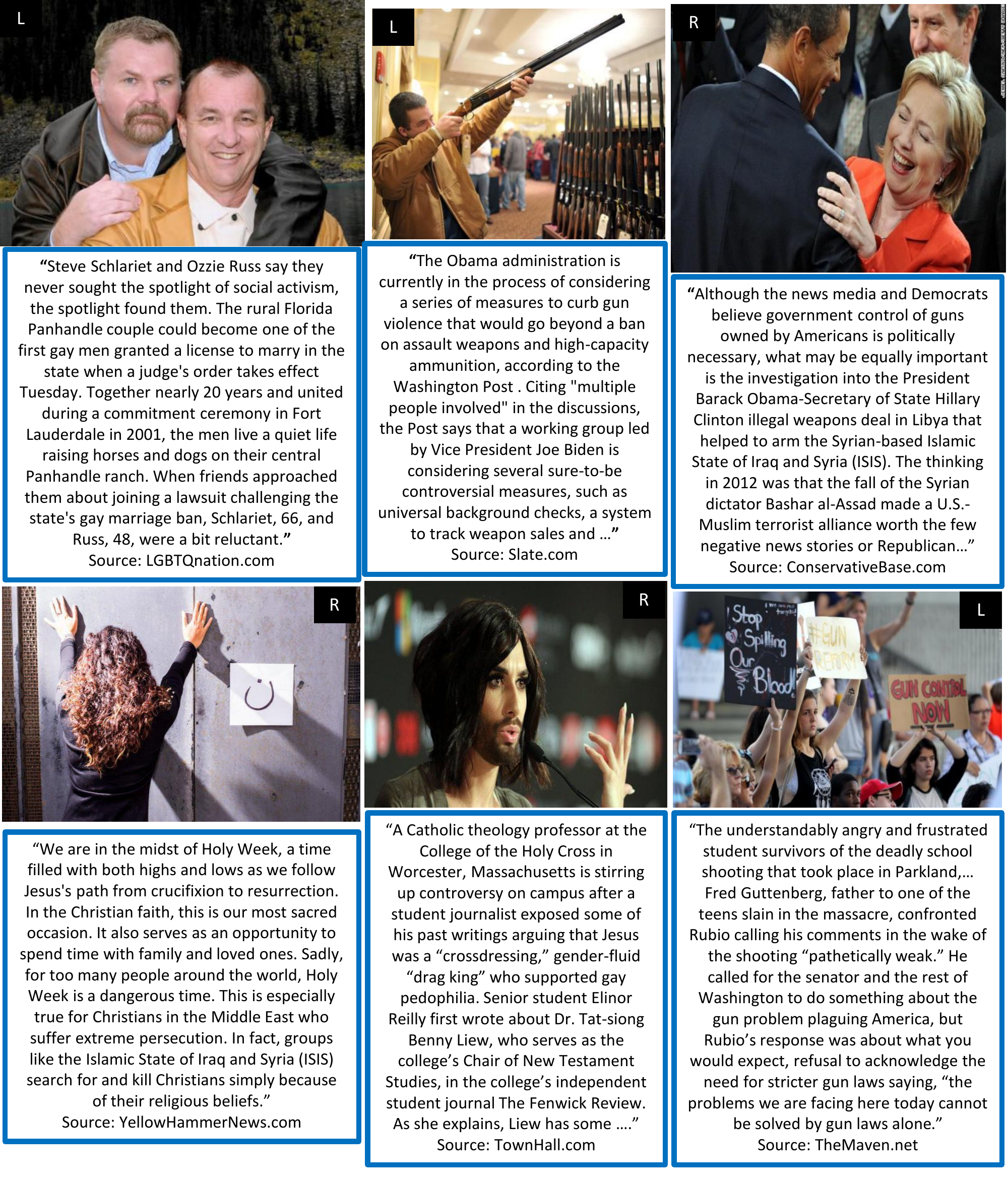}
    \caption{Example images and articles (truncated) from our dataset. We annotate each image with the media source from which it and the article came, as well as the politics of that media source, as determined by Media Bias Fact Check (see our main text for details).}
    \label{fig:dataset_ex3}
\end{figure}

In Figures \ref{fig:dataset_ex1}-\ref{fig:dataset_ex3}, we show example images and some text from their associated articles from our dataset. Note that the text we include for each image is truncated, as many of the articles are quite lengthy. We also annotate each image with a ``L'' or ``R'' depending on whether the image comes from the left or right respectively, as well as the original source for the image and article text. 

We believe these images highlight how extreme some of our media sources are. For example, in Fig.\ \ref{fig:dataset_ex1}, we see an image of apparent Hispanic gang members with Obama's head cropped as one of them. The article discusses illegal immigration and alleges Obama has facilitated allowing ``illegals'' to stay. In Fig.\ \ref{fig:dataset_ex2}, we see several images of protests, one of which is associated with the left and one of which is associated with the right. We note, however, that the protest image associated with the right (bottom left) actually shows protesters carrying signed \textit{supportive} of Planned Parenthood, a topic associated with the left. A similar situation is found in Fig.\ \ref{fig:dataset_ex3} where we see an image of a transgender man (bottom row, middle) labeled right. These images' labels only makes sense in conjunction with their paired article text, which are describing circumstances related to the image. These examples underscore one primary challenge of learning visual classifiers on our dataset: \textit{images' labels often depend upon the context on which they appear as much as they depend on what is in the image itself. } Thus, learning to predict the politics from an image is highly challenging due to the inherent high-level semantic nature of the problem as well as the presence of noisy data. We believe our method, guided by privileged information from the text domain helps guide the training and alleviates some of these problems.

\null
\vfill

\null
\vfill

\null
\vfill

\null
\vfill

\null
\vfill

\null
\vfill

{\small
\bibliographystyle{ieee}
\bibliography{supplementary}
}